\documentclass[conference]{IEEEtran}
\IEEEoverridecommandlockouts

\usepackage{amsmath,amssymb,amsfonts}
\usepackage{algorithmic}
\usepackage{graphicx}
\usepackage{textcomp}
\usepackage{xcolor}
\usepackage{siunitx}
\usepackage{xcolor}
\usepackage{siunitx}
\usepackage{hyperref}
\usepackage{siunitx}
\usepackage{caption}
\usepackage{wrapfig}
\usepackage[list=true]{subcaption}

\begin{document}

\title{GenAI for Energy-Efficient and Interference-Aware Compressed Sensing of GNSS Signals\\on a Google Edge TPU\\
{\footnotesize \thanks{This work has been carried out within the DARCII project, funding code 50NA2401, sponsored by the German Federal Ministry for Economic Affairs and Climate Action (BMWK) and supported by the German Aerospace Center (DLR), the Bundesnetzagentur (BNetzA), and the Federal Agency for Cartography and Geodesy (BKG).}}}

\author{\IEEEauthorblockN{Thorben Wegner\IEEEauthorrefmark{1},
    Lucas Heublein\IEEEauthorrefmark{1},
    Tobias Feigl\IEEEauthorrefmark{1}\IEEEauthorrefmark{2},
    Felix Ott\IEEEauthorrefmark{1},
    Christopher Mutschler\IEEEauthorrefmark{1},
    Alexander Rügamer\IEEEauthorrefmark{1}
  }
  \IEEEauthorblockA{\IEEEauthorrefmark{1}Fraunhofer Institute for Integrated Circuits IIS, 90411 Nürnberg, Germany}
  \IEEEauthorblockA{\IEEEauthorrefmark{2}Friedrich-Alexander-Universität Erlangen-Nürnberg, 91058 Erlangen, Germany}
  \IEEEauthorblockA{\{lucas.heublein, tobias.feigl, felix.ott, christopher.mutschler, alexander.ruegamer\}@iis.fraunhofer.de}
}
\maketitle
\begin{abstract}
Traditional methods for classifying global navigation satellite system (GNSS) jamming signals typically involve post-processing raw or spectral data streams, requiring complex and costly data transmission to cloud-based interference classification systems. In contrast, our proposed approach efficiently compresses GNSS data streams directly at the hardware receiver while simultaneously classifying jamming and spoofing attacks in real time. Given the growing prevalence of GNSS jamming, there is a critical need for real-time solutions suitable for power-constrained environments. 
This paper introduces a novel method for compressing and classifying GNSS jamming threats using generative artificial intelligence (GenAI), specifically variational autoencoders (VAEs), deployed on Google Edge tensor processing units (TPUs). The study evaluates various autoencoder (AE) architectures to compress and reconstruct GNSS signals, focusing on preserving interference characteristics while minimizing data size near the receiver hardware. The pipeline adapts large-scale AE models for Google Edge TPUs through 8-bit quantization to ensure energy-efficient deployment. Tests on raw in-phase and quadrature-phase (IQ) data, Fast Fourier Transform (FFT) data, and handcrafted features show the system achieves significant compression ($>42\times$) and accurate classification of approximately 72 interference types on reconstructed signals ($\text{F}_{2}$-score 0.915), closely matching the original signals ($\text{F}_{2}$-score 0.923). The hardware-centric GenAI approach also substantially reduces jammer signal transmission costs, offering a practical solution for interference mitigation. Ablation studies on conditional and factorized VAEs (i.e., FactorVAE) explore latent feature disentanglement for data generation, enhancing model interpretability and fostering trust in machine learning (ML) solutions for sensitive interference applications.
\end{abstract}

\section{Introduction}

Interference signals adversely affect the processing chain of GNSS, thereby degrading or completely inhibiting GNSS-based localization capabilities~\cite{Ioannides2016Vulnerabilities,merwe_franco}. To ensure the continued functionality of localization systems, it is imperative to eliminate potential interference sources, including intentional jammers~\cite{heublein_raichur_ion,ott_heublein_icl}. This objective necessitates the detection, classification, and localization of jammers, requiring the development of novel systems~\cite{Merwe2022,gaikwad_heublein} capable of collecting detailed data to monitor communication channels of GNSS receivers~\cite{heublein_feigl_crpa,manjunath_heublein,heublein_feigl_posnav}.

The granularity of the data collected plays a crucial role in accurately representing potential interference. However, electromagnetic field (EMF) sensors designed for spectral monitoring of GNSS signals generate extensive data, typically at rates around 2 to 60\,MHz. For practical deployment, it is essential to minimize the volume of data transmitted from the GNSS sensor to remote analysis systems. Large data volumes not only increase power consumption but also significantly raise operational costs, rendering real-world applications infeasible. For instance, an interference detector without data compression transmits approximately 346\,GB of data per day (at $4\frac{\text{MB}}{s}$) over cellular networks, incurring daily costs of approximately 900\,USD (=$346\cdot\frac{2.6\$}{\text{GB}}$) in Germany~\cite{WorldPrice2022}. Consequently, (IOT crowdsourcing) interference detectors without compression mechanisms are not economically viable. 

While it is necessary to compress fine-grained data, conventional compression techniques~\cite{Libutti2020BenchmarkingPA} often reduce the resolution of interference signals, diminishing their informational value. A reduction in signal representation quality compromises the effectiveness of detection and classification systems, thereby impeding the localization and suppression of interference sources. Although advanced compression algorithms~\cite{ComparisonLossKodituwakku} can preserve more information, their computational complexity increases system costs and makes real-time deployment impractical. Recent ML approaches have demonstrated the potential for accurate and robust jammer classification, even in real-world conditions~\cite{heublein_raichur_ion}. However, these methods typically rely on large-scale models and extensive datasets, necessitating significant computational resources or cloud-based infrastructure. As a result, the challenge remains: how to develop an innovative anti-jammer system that efficiently compresses interference signals while maintaining cost- and energy-efficiency, without compromising essential signal characteristics for effective detection and classification.

\vspace{+0.1cm}
\textbf{Contributions.} This paper addresses the challenge of efficiently compressing the vast amounts of data generated by mobile GNSS interference detectors while preserving critical information for identifying interference sources. We present an in-depth investigation of various state-of-the-art GenAI models, including AEs, vanilla VAEs, conditional VAEs, and factorized VAEs~\cite{kim_mnih}. These models are trained on diverse GNSS data representations, encompassing raw in-phase/quadrature (IQ) data, FFT spectrum data, and pre-processed hand-crafted features such as kurtosis, entropy, spectral energy, and their combinations. Our central hypothesis posits that integrating spectral and temporal information enables the capture of complementary features, leading to more efficient and characteristic data compression. We evaluate these models across multiple performance dimensions to identify architectures that achieve the highest compression rates, lowest power consumption, and most accurate signal reconstructions for downstream interference detection and classification tasks. To rigorously assess feature reconstruction accuracy, we develop a downstream classification pipeline based on an ensemble Random Forest classifier. This pipeline evaluates classification accuracy and uncertainty across different input formats and (V)AE architectures. Additionally, we analyze the combinations of features -- such as IQ, spectrum, and temporal data -- that maximize both reconstruction and classification performance. The classification task covers five interference categories (continuous wave, chirp, noise, pulsed, and broadband interference), comprising 72 specific classes, using original, reconstructed, and latent data representations. Following the identification of the most efficient and accurate processing pipeline, we apply advanced compression techniques, including 8-bit quantization, to further optimize the models. The (V)AE-based pipeline is converted from PyTorch to TensorFlow and adapted for deployment on resource-constrained microcontrollers, specifically Google Edge TPUs such as Google Coral~\cite{yazdanbakhsh2021evaluation}. We comprehensively benchmark the compressed processing pipeline across several key metrics, including power efficiency (in watts), compressibility (latent variables), reconstruction accuracy, downstream classification performance, energy consumption, data transmission rates, and detection/classification accuracy ($\text{F}_{2}$-score). The evaluation results demonstrate the practical applicability of our approach, highlighting its potential for efficient and accurate real-world GNSS interference detection and classification.

\vspace{+0.1cm}
\textbf{Outline.} Section~\ref{sec:related_work} provides an extensive review of relevant literature. Section~\ref{label_background} outlines the background on wave signals and compression. The proposed methodology and preprocessing pipeline are described in Section~\ref{sec:methodology}. Details of the experimental setup are presented in Section~\ref{sec:experiments}, followed by a presentation and discussion of the results in Section~\ref{sec:results}. Finally, Section~\ref{sec:summary} concludes.
\section{Related Work}
\label{sec:related_work}

State-of-the-art methods~\cite{ComparisonTimeSeriesFreeman} apply FFT~\cite{Duhamel1990} to raw IQ data from GNSS sensors for compression and the extraction of hand-crafted features. Experts then manually select features, such as entropy, spectrum, and kurtosis, based on their relevance to detecting and classifying potential jammers. However, these compression techniques are lossy, as the spectrum fails to preserve time information, thereby losing important details that could enhance detection and classification accuracy. While FFT-based compression is computationally efficient, it produces a memory-intensive feature set that increases the data transmission rate, making it impractical for mobile interference analysis systems (MIAS). Moreover, Google's Coral low-power TPUs, commonly used for detection and classification tasks, do not provide an FFT implementation. As a result, promising new compression algorithms capable of detecting, classifying, and directly locating jammers in raw IQ data from GNSS sensors remain largely unexplored. In contrast, research in other domains, such as image and language processing, demonstrates that generative neural networks, including AEs, can efficiently compress large data streams~\cite{Govorkova_2022} while simultaneously detecting anomalies in a problem-specific manner~\cite{Abbasi2021Outlier}. However, it remains unclear whether and how these methods can be adapted to GNSS data, and how reliably and accurately they can detect and classify anomalies in this context. Additionally, it is uncertain whether such methods can be implemented in an energy-efficient manner on integrated circuits (e.g., TPUs) and whether they can significantly reduce data traffic, making MIAS more practical. Furthermore, prior compression algorithms have not integrated techniques such as pruning~\cite{Blalock2020Pruning}, quantization~\cite{Yang2019Quantization}, or Huffman coding~\cite{Xu2022SurvCom} to further minimize the energy footprint.

Another research effort, proposed by Du et al.~\cite{du_min_guo} applies a self-supervised denoising convolutional AE (DCAE) to compress Delay/Doppler Maps (DDMs) in GNSS reflectometry. This method effectively reduces data volume by approximately 90\% while preserving essential features for subsequent analysis. The encoder-decoder structure of the DCAE demonstrates its capability to compress and restore DDMs, highlighting the potential of disentanglement techniques in GNSS data compression. Another approach presented by Ansari et al.~\cite{ansari_kaushik} utilizes compressive sensing (CS) theory to compress and reduce the size of raw GNSS data intended for transmission. This method aims to efficiently transmit data for further processing, addressing the challenges associated with large data volumes in GNSS applications. Li et al.~\cite{li_wang_zhong} introduce the FD-DEFLATE compression scheme, offering a promising solution for efficient GNSS data compression in the remote transmission of GNSS interference. This scheme aims to enhance the efficiency of data transmission in GNSS systems. However, research on compression methods that leverage disentanglement remains largely unexplored.

\begin{figure*}[!t]
    \centering
	\begin{minipage}[t]{0.325\linewidth}
        \centering
    	\includegraphics[width=0.64\linewidth]{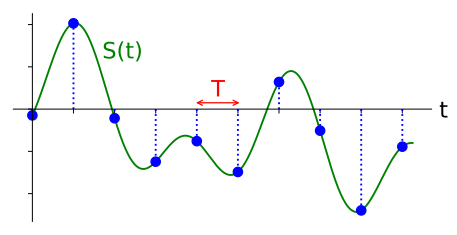}
        \subcaption{Continuous signal with interval $T$.}
        \label{fig:ImSampling}
    \end{minipage}
    \hfill
	\begin{minipage}[t]{0.325\linewidth}
        \centering
    	\includegraphics[width=1.0\linewidth]{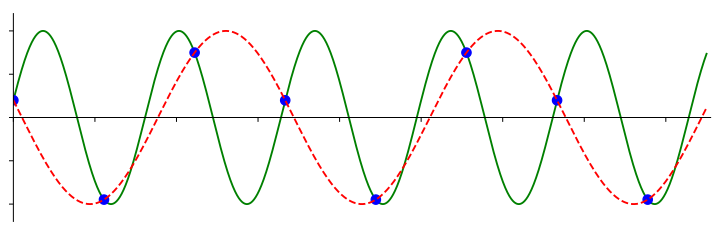}
        \subcaption{Sampling with sampling rate $f_s = 1.5f$.}
        \label{fig:NyquistHalf}
    \end{minipage}
    \hfill
	\begin{minipage}[t]{0.325\linewidth}
        \centering
    	\includegraphics[width=1.0\linewidth]{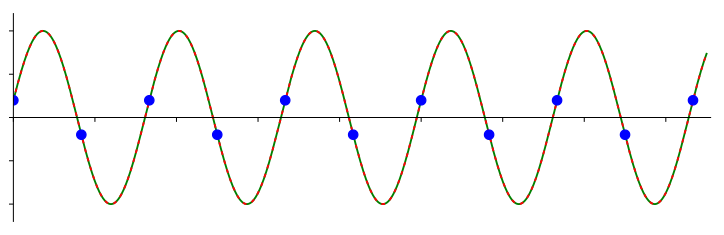}
        \subcaption{Sampling with sampling rate $f_s = 2f$.}
        \label{fig:NyquistFull}
    \end{minipage}
    \caption{Various signals with different sampling intervals/rates~\cite{PySDR}.}
    \vspace{-0.25cm}
\end{figure*}

There is a substantial body of research on GNSS interference monitoring using ML. However, this research has not yet explored the context of compression. Shen et al.~\cite{shen_chen_wang} employed ML models to predict and correct GNSS errors. Brieger et al.~\cite{Brieger2022Classification} took into account both the spatial and temporal relationships between samples when fusing data from a coarse-grained low-resolution LC sensor and data from a bandwidth-limited LC sensor. Jia et al.~\cite{jia_zhang} proposed a multimodal feature fusion network (MFFNet) that integrates multiple signal features to accurately identify GNSS interference types, achieving over 92\% accuracy even for low-power signals. Khan et al.~\cite{khan_aguado} introduced a hybrid jammer frequency estimation method, combining spectrogram and adaptive notch filter approaches, which enabled superior interference suppression in GNSS signals via zero-phase notch filtering. Raichur et al.~\cite{raichur_ion_gnss} and Borio et al.~\cite{borio_gioia_stern} proposed a crowdsourcing approach using smartphone-based features to localize the source of detected interference. Raichur et al.~\cite{raichur_heublein} adapted to novel interference classes through continual learning, specifically employing class-incremental learning techniques. Heublein et al.~\cite{heublein_feigl_crpa} introduced an extensive dataset of snapshots obtained from a low-frequency antenna, capturing a wide range of generated interferences within a large-scale environment, including controlled multipath effects. Mehr et al.~\cite{mehr_savolainen} proposed a dual-stage architecture consisting of an LSTM-based AE. Liu et al.~\cite{liu_han} introduced a meta-learning-based GNSS interference recognition scheme, using MAML+SE-ResNet to enhance generalization and accuracy in few-shot scenarios.
\section{Background}
\label{label_background}

This section provides the foundational background necessary to implement the proposed compression pipeline, including the principles underlying the digitization of continuous wave signals such as GNSS. Key concepts such as IQ sampling, Nyquist theory, and quadrature sampling are introduced (see Section~\ref{IQData}). Section~\ref{AutoEncoder} presents a concise overview of AE architectures. A brief overview of the DevBoard Mini is provided in Section~\ref{DevBoardMini}. Finally, Section~\ref{label_languages} outlines the programming languages employed. The ONNX format, used as an intermediate step for converting models from PyTorch to TensorFlow Lite, is discussed in Section~\ref{ONNXFormat}.

\subsection{IQ, Nyquist \& Quadrature Sampling}
\label{IQData}

\textit{Sampling} refers to the process of converting a continuous-time signal into a discrete-time signal. This involves recording the continuous signal $S(t)$ at uniform intervals of $T$ seconds, where $T$ is referred to as the sampling period or sampling interval. The sampling frequency, defined as $\frac{1}{T}$ or $f_s$, represents the rate at which samples are taken and is measured in Hertz (Hz) or $\frac{1}{s}$. An illustration of this process is provided in Figure~\ref{fig:ImSampling}~\cite{PySDR}. In signal processing, the \textit{Nyquist frequency} for a given sampling rate is defined as the frequency whose period is twice the sampling interval, corresponding to $0.5\ \frac{\text{cycles}}{\text{sample}}$. This implies that the sampling rate must be at least twice the highest frequency present in the sampled signal, or $f_s = 2.0f$. When the highest frequency of a signal is less than the Nyquist frequency of a given sampling rate, the resulting discrete-time sequence remains free of distortion. In this case, the sampling rate is considered to be above the Nyquist rate for that particular signal~\cite{PySDR}. To illustrate the significance of the Nyquist rate, consider a simple sine wave with frequency $f$, as shown by the green line in Figure~\ref{fig:NyquistHalf}. If the signal is sampled at a rate of $f_s = 1.5f$, the resulting reconstruction, indicated by the red dashed line, intersects each sample point. This reconstructed sine wave has the lowest possible frequency that still intersects the sampled points. However, since the sampling rate is insufficient, the reconstructed signal fails to accurately represent the original sine wave. Conversely, when the sampling rate is $f_s = 2.0f$, which corresponds to the Nyquist rate (Figure~\ref{fig:NyquistFull}), the reconstructed sine wave matches the frequency of the original signal exactly.

The term \textit{quadrature} refers to two waves that are 90 degrees out of phase with one another. Since cosine and sine waves are inherently 90 degrees out of phase, they are used to represent the sampled signal in \textit{quadrature sampling}. We introduce the variables $I$ and $Q$ to represent the amplitudes of the cosine (in-phase) and sine (quadrature) components, respectively. Specifically, the cosine wave represents the \textit{in-phase} component, denoted as $Icos(2\pi ft)$, and the sine wave represents the \textit{quadrature} component, denoted as $Qsin(2\pi ft)$. The signal is then represented as $x(t)=Icos(2\pi ft) + Qsin(2\pi ft)$. A slight change in either component $I$ or $Q$ results in a corresponding slight change in the amplitude and phase of the resulting signal. This relationship can be derived from the trigonometric identity: $acos(x) + bsin(x)=Acos(x-\phi)$. The advantage of this identity is that the amplitude $A$ and phase $\phi$ of the resulting sine wave can be controlled by adjusting the amplitudes $I$ and $Q$. This approach allows for phase manipulation by varying the amplitudes, which is simpler than directly altering the phase of the input waves.

\subsection{Autoencoders}
\label{AutoEncoder}

\textbf{Autoencoders} (AEs)~\cite{Bank2020AE} are a class of neural networks used for unsupervised learning, data compression, and denoising. They function by encoding input data into coordinates within a latent space, which often has a dimensionality reduced by a factor of ten or more compared to the original input. This latent representation is then decoded to reconstruct data that closely approximates the original input. An AE consists of two primary components: an encoder and a decoder. The encoder maps the input to a lower-dimensional latent representation, while the decoder reconstructs the input from this latent representation. During training, the AE is optimized to minimize the reconstruction error between the original input and its reconstruction, enabling it to effectively learn meaningful transformations to and from the latent space.

\vspace{+0.1cm}
\textbf{Convolutional Autoencoders} (CAEs) are a specialized variant of AEs particularly well-suited for image data, though they can also process one-dimensional data. Similar to traditional AEs, CAEs comprise an encoder and a decoder. However, in CAEs, these components are constructed using CNNs~\cite{CNNandAE,DeppLearningBook}, which are specifically designed for efficient image processing. During training, the CAE is optimized to minimize the reconstruction error between the original input and the reconstructed output, enabling it to effectively capture and represent essential features of the data.

\vspace{+0.1cm}
\textbf{Variational Autoencoders} (VAEs)~\cite{Variational_AE} are a class of generative models that utilize neural networks to learn the underlying structure of a dataset and generate new samples resembling the original data. Unlike traditional AEs, VAEs capture a probabilistic representation of the latent space, enabling the generation of novel, previously unseen samples. In addition to minimizing reconstruction error, VAEs optimize a loss function that encourages the latent space to follow a predefined probability distribution, typically a normal distribution. This is achieved by incorporating a regularization term, which penalizes deviations between the learned latent space distribution and the target distribution. The regularization promotes a smooth and continuous latent space, facilitating the interpolation and generation of new, coherent samples.

\subsection{Google Coral Tensor Processing Units}
\label{DevBoardMini}

\setlength{\intextsep}{6pt}
\setlength{\columnsep}{12pt}
\begin{wrapfigure}{R}{4.7cm}
    \begin{minipage}[b]{1.0\linewidth}
        \centering
        \includegraphics[trim=370 160 450 160, clip, width=1.0\linewidth]{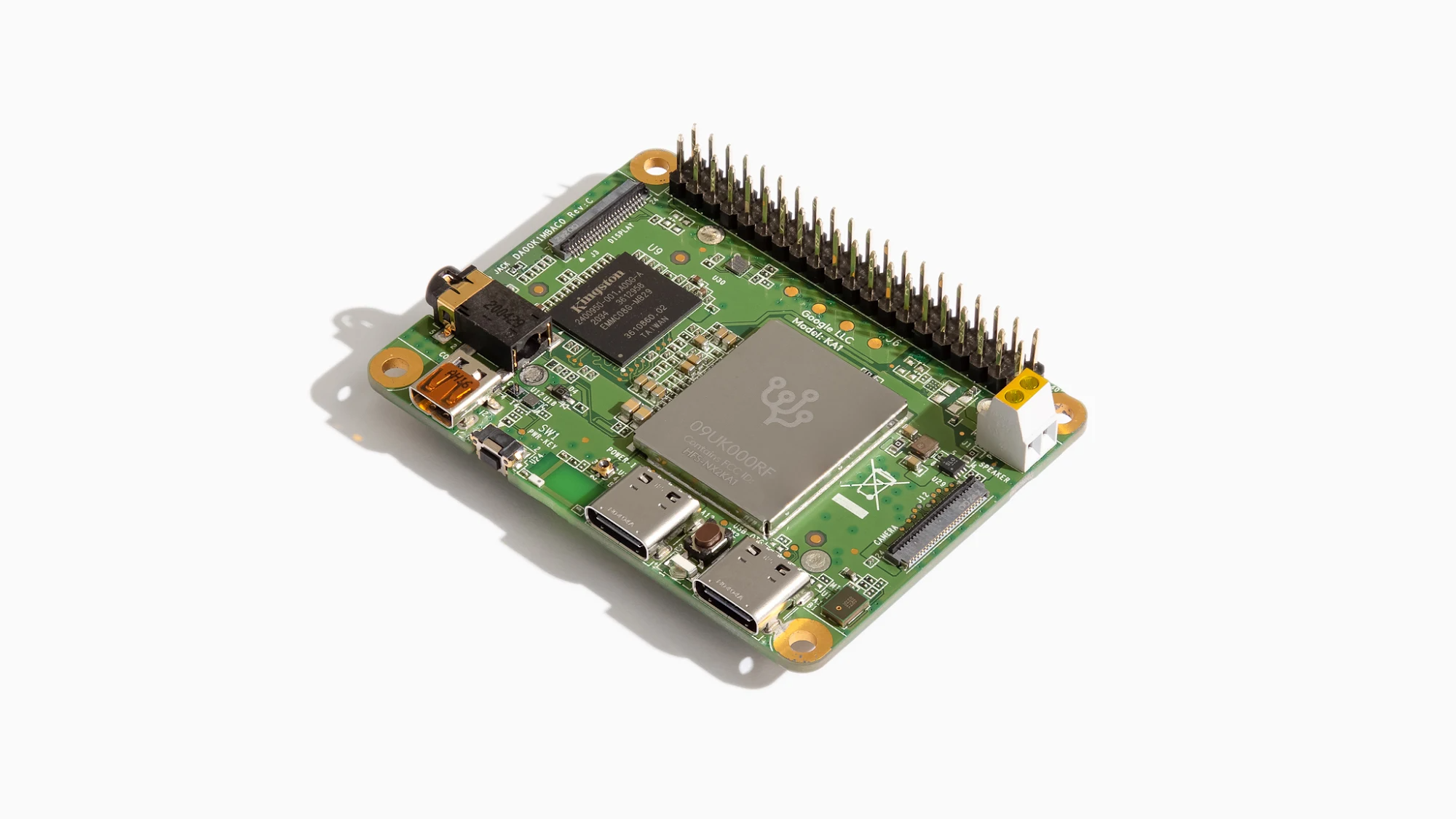}
        \caption{Image of the Coral DevBoard Mini~\cite{DevBoardMini}.}
        \label{fig:ImDevBoardMini}
    \end{minipage}
\end{wrapfigure}
The Coral DevBoard Mini~\cite{DevBoardMini} (see Figure~\ref{fig:ImDevBoardMini}) is a single-board computer designed for fast ML inference. It primarily functions as an evaluation platform for the Accelerator Module, a surface-mounted module that integrates the Edge tensor processing unit (TPU). The Edge TPU, which has a peak performance of 4 TOPS with \textit{int8} precision, operates at a power consumption of 2\,W, delivering 2 TOPS per watt~\cite{DevBoardMiniDatasheet}. This small ASIC, developed by Google, accelerates TensorFlow Lite models. The Edge TPU is connected to the main system-on-chip (SoC) via USB 2.0. The DevBoard Mini is equipped with a total of 256\,KB + 64\,KB of RAM. Due to the limited memory capacity, the models deployed on the device must be compact.

\subsection{Programming Languages}
\label{label_languages}

\textit{PyTorch} is an open-source ML library developed by Facebook's AI research team, renowned for its dynamic computational graph. This feature enables users to modify and inspect models during both training and inference, offering significant flexibility. Such flexibility facilitates the implementation of complex models and encourages innovative research. PyTorch also provides a comprehensive suite of tools for constructing neural networks, supports data processing, and utilizes GPU acceleration to optimize training and inference efficiency.

\textit{TensorFlow} is an open-source deep learning framework developed by Google, designed for constructing and deploying ML applications using data flow graphs that can be executed across GPUs, CPUs, and TPUs on various platforms. It supports both computational graphs for efficient model training and execution for immediate evaluation of operations. Keras serves as a high-level API and TensorFlow streamlines model development and offers scalability, automatic gradient computation, and cross-platform interoperability.

\textit{TensorFlow Lite} is a streamlined version of TensorFlow, specifically designed for mobile and embedded devices. It facilitates the deployment of trained TensorFlow models on mobile platforms, such as smartphones and tablets, as well as on embedded systems, including microcontrollers and sensors. Given the limited memory of microcontrollers and embedded devices, which often support only lower precision formats like \textit{int8} or \textit{float16}, whereas TensorFlow models are typically trained using 32-bit floating-point numbers, TensorFlow Lite offers the ability to adjust precision to \textit{float16}, \textit{bfloat16}, or \textit{int8} through model optimization techniques. These optimizations reduce memory usage and computational requirements while preserving accuracy. Key techniques include model quantization, which reduces the precision of weights and activations, and model pruning, which removes redundant components.

\subsection{The ONNX Format}
\label{ONNXFormat}

The ONNX format~\cite{ONNXGit} serves as an intermediate representation for storing and communicating various types of neural networks. In the ONNX format, neural networks are typically represented as graphs, commonly referred to as ONNX graphs. These graphs consist of nodes, also known as layers or modules, which represent standard operators. The nodes are connected by edges that define the data flow through the computational graph. Inference on such graphs is performed by tracing data flows from the input to the output nodes, queuing computations accordingly. The ONNX format provides an abstract intermediate representation of the compute graphs of neural networks. Many deep learning frameworks now support exporting models to the ONNX format; for instance, PyTorch includes the \textit{torch.onnx} module for exporting PyTorch models to ONNX. In our work, we use the ONNX format as an intermediary step during the conversion process from PyTorch to TensorFlow. All nodes within a graph are stored in topological order, with each node representing a single mathematical operation encoded using a type string. Each node contains sets of input edges, output edges, and parameters. Data in ONNX is represented using tensors, which are not stored within the nodes themselves but are instead kept by the graph in an unordered dictionary. Nodes perform their operations by utilizing their assigned input edges, parameters, and pre-initialized tensors, and subsequently pass the results to their designated output edges.
\section{Methodology}
\label{sec:methodology}

The primary objective of this paper is to explore the efficient compression of GNSS signals through the application of GenAI techniques, particularly VAEs, with a focus on their deployment on Edge TPUs. This investigation involves the evaluation of various AE architectures, including fully connected, convolutional, and variational AEs, to identify the most accurate and memory-efficient approaches for signal representation while minimizing reconstruction errors. The research includes the development of several (V)AE architectures, both in their standard form and in an 8-bit compressed variant, emphasizing reconstruction accuracy, classification performance (measured by the $\text{F}_{2}$-score), and energy efficiency. To examine the impact of compression on classification accuracy, a downstream classification pipeline is implemented using random forests (RFs) applied to the reconstructed signals, with performance visualized through confusion matrices.

First, we describe the structure of the various pipelines deployed, including the classification pipeline, which processes different types of input data, performs reconstruction, and conducts classification (refer to Section~\ref{simpleclassification}). Additionally, details are provided on the conversion of PyTorch models to TensorFlow Lite models, enabling their compilation and execution on the Coral Edge TPU via the ONNX format. In Section~\ref{FindingBestArchitecture}, we outline the process for identifying the optimal model architectures for each input type. Finally, we present our disentanglement approach for compressing GNSS signals using FactorVAE (see Section~\ref{label_disentanglement}).

\subsection{Pipeline}
\label{simpleclassification}

\textbf{Overview.} The pipeline proposed in this paper offers a structured and systematic approach for the compression and classification of GNSS signals using (V)AEs, optimized specifically for deployment on Edge TPUs. The process begins with the training of various (V)AE architectures on diverse datasets, including temporal, spectral, and mixed-domain data. Initially, simpler datasets are used to ensure adequate reconstruction accuracy, with progressively more complex data introduced as training advances. Upon completion of training, the (V)AEs are converted from PyTorch to TensorFlow Lite to ensure compatibility with Edge TPUs. This conversion involves an intermediate step using the ONNX format, followed by compilation via the Edge TPU compiler. To evaluate the effect of compression on classification accuracy, a downstream classification pipeline employing RFs is implemented. Reconstruction errors and classification results are visualized through confusion matrices, offering valuable insights into the performance of the models. The pipeline demonstrates an efficient methodology for compressing and classifying GNSS signals, balancing reconstruction accuracy and classification performance for real-world applications. Given that the primary objective is to identify the most accurate and memory-efficient compression techniques for GNSS signals using AEs on Edge TPUs, rather than developing the most optimized classification pipeline, a simple proof-of-concept classification approach is employed to validate the real-world applicability of the proposed framework. For achieving optimal classification performance in practical scenarios, adopting a more sophisticated approach, such as that presented in \cite{Brieger2022Classification}, is recommended for downstream classification tasks.

\vspace{+0.1cm}
\textbf{Classification Task.} For the classification task, three distinct RF models~\cite{fawagreh2014random} are trained, each comprising 7,000 estimators. This setup is applied to three different input types: temporal domain data, spectral domain data, and a combination of both. The first RF is trained on raw, uncompressed data to establish a baseline classification accuracy, which serves as a reference for evaluating the impact of compression. This baseline allows for assessing the loss in classification accuracy resulting from the compression of raw input data using an AE, as well as the additional accuracy loss due to the compression and quantization of the AE itself, which affects reconstruction accuracy. The second RF is trained on the reconstructed data output by an uncompressed and unquantized AE. The third model is trained on data reconstructed by the compressed and quantized AE deployed on the Edge TPU.

\vspace{+0.1cm}
\textbf{From PyTorch to Edge TPU.} All AEs for the different input types are trained using the PyTorch deep learning framework. As a result, the models must be converted to TensorFlow Lite for compatibility with the Edge TPU on the DevBoard Mini. This conversion process is accomplished using the Edge TPU Compiler~\cite{EdgeTPUCompiler} and involves four distinct steps. In the first step, the trained AE is converted to the ONNX format using the \textit{torch.onnx} package~\cite{TorchONNXExport}, which transforms the PyTorch model into an ONNX graph. The second step involves converting the ONNX graph to a TensorFlow Graph~\cite{tfGraph}. This is facilitated by the TensorFlow Backend for ONNX, which allows ONNX models to be used as input for TensorFlow. The model is initially converted to a TensorFlow format and subsequently executed on TensorFlow to generate the output~\cite{ONNXTensorFlowGit}. The third step of the pipeline converts the TensorFlow graph model to TensorFlow Lite, applying post-training integer quantization to 8-bit format. This quantization is necessary since the Edge TPU on the DevBoard Mini only supports 8-bit integer inputs. In the fourth step, we compile the TensorFlow Lite model for deployment on the Edge TPU.

\subsection{Architectures}
\label{FindingBestArchitecture}

Next, we present a description of the various architectures designed for different data types, including spectral data (Section~\ref{trainFrequency}), raw IQ data (Section~\ref{label_raw_iq_data}), temporal data (Section~\ref{label_temporal_data}), and mixed spectral-temporal data (Section~\ref{label_spectral_temporal_data}).

\vspace{+0.1cm}
\textbf{Spectral Domain Data.}
\label{trainFrequency}
\begin{figure}[!t]
    \centering
    \includegraphics[width=0.712\linewidth]{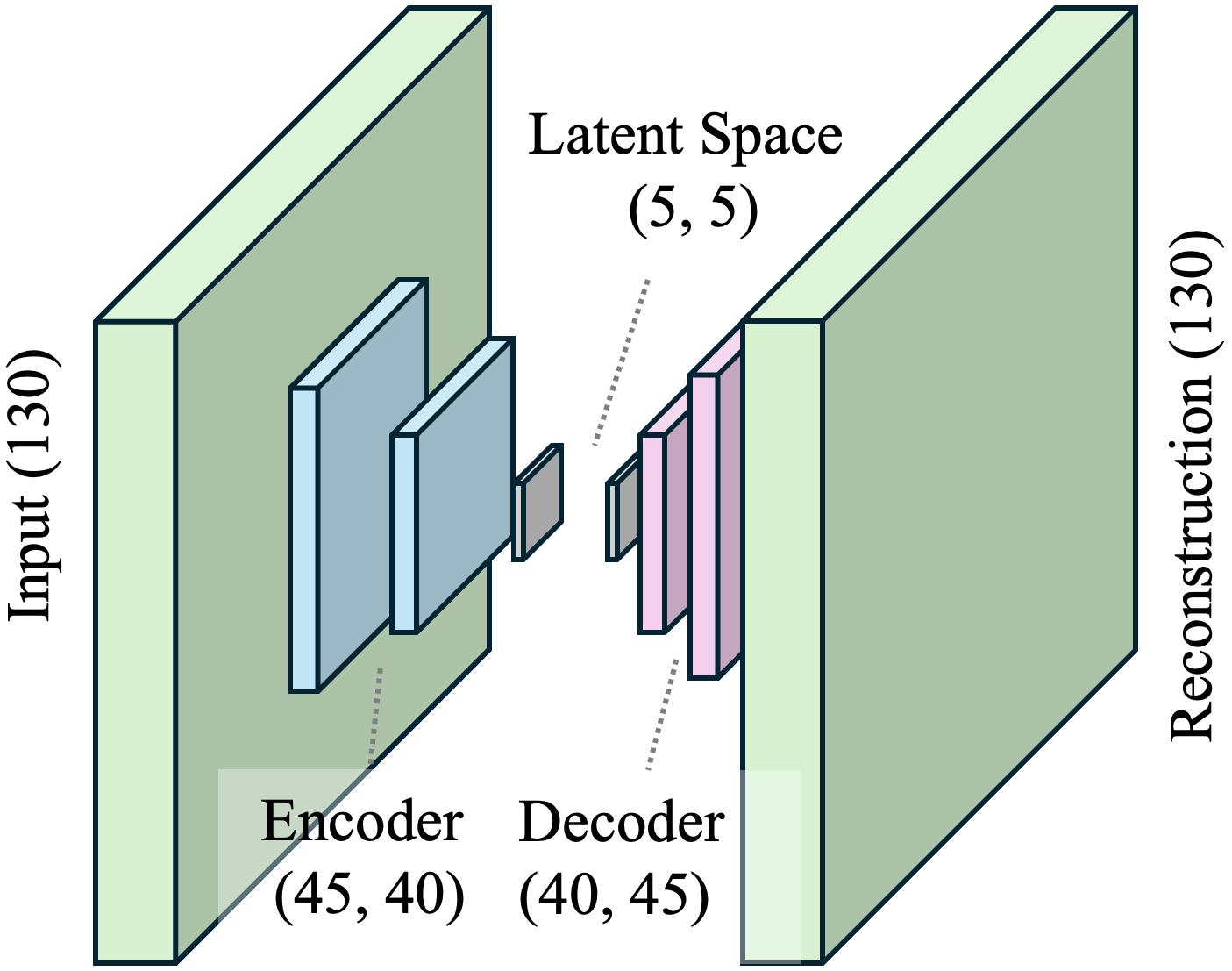}
    \caption{Our fully connected AE comprising 2 hidden layers.}
    \label{fig:AEArch:SDR3}
\end{figure}
For the spectral domain, a fully connected AE is utilized, with both the encoder and decoder consisting exclusively of fully connected layers. Data for the spectral domain comprises four distinct scenarios. In the initial step, the AEs are trained using data from the first scenario, which contains the simplest data and is therefore easier for the models to learn. If acceptable levels of reconstruction and classification accuracy are achieved for this scenario, the process is repeated using data from all four scenarios combined. The optimal architecture for each of these steps is identified using a consistent workflow. All reasonable architectures are initially trained for 90 epochs. Various combinations of the number of hidden layers, the number of neurons per hidden layer, and the latent space dimensions are explored. The number of hidden layers ranges from two to three, and the number of neurons per layer can be 32, 64, or 128. However, a deeper layer may not have more neurons than a preceding shallower layer. For instance, if the first hidden layer has 64 neurons, the second hidden layer can only contain 32 or 64 neurons, but not 128. The number of latent variables ranges from three to ten, resulting in a total of 128 possible combinations. An example architecture is illustrated in Figure~\ref{fig:AEArch:SDR3}. Upon completing the initial training of the 128 AEs, the 14 architectures with the lowest reconstruction error are selected for further training. These selected models are retrained for max.~700 epochs (early stopping) to refine their performance.

\begin{figure}[!t]
    \centering
    \includegraphics[width=1.0\linewidth]{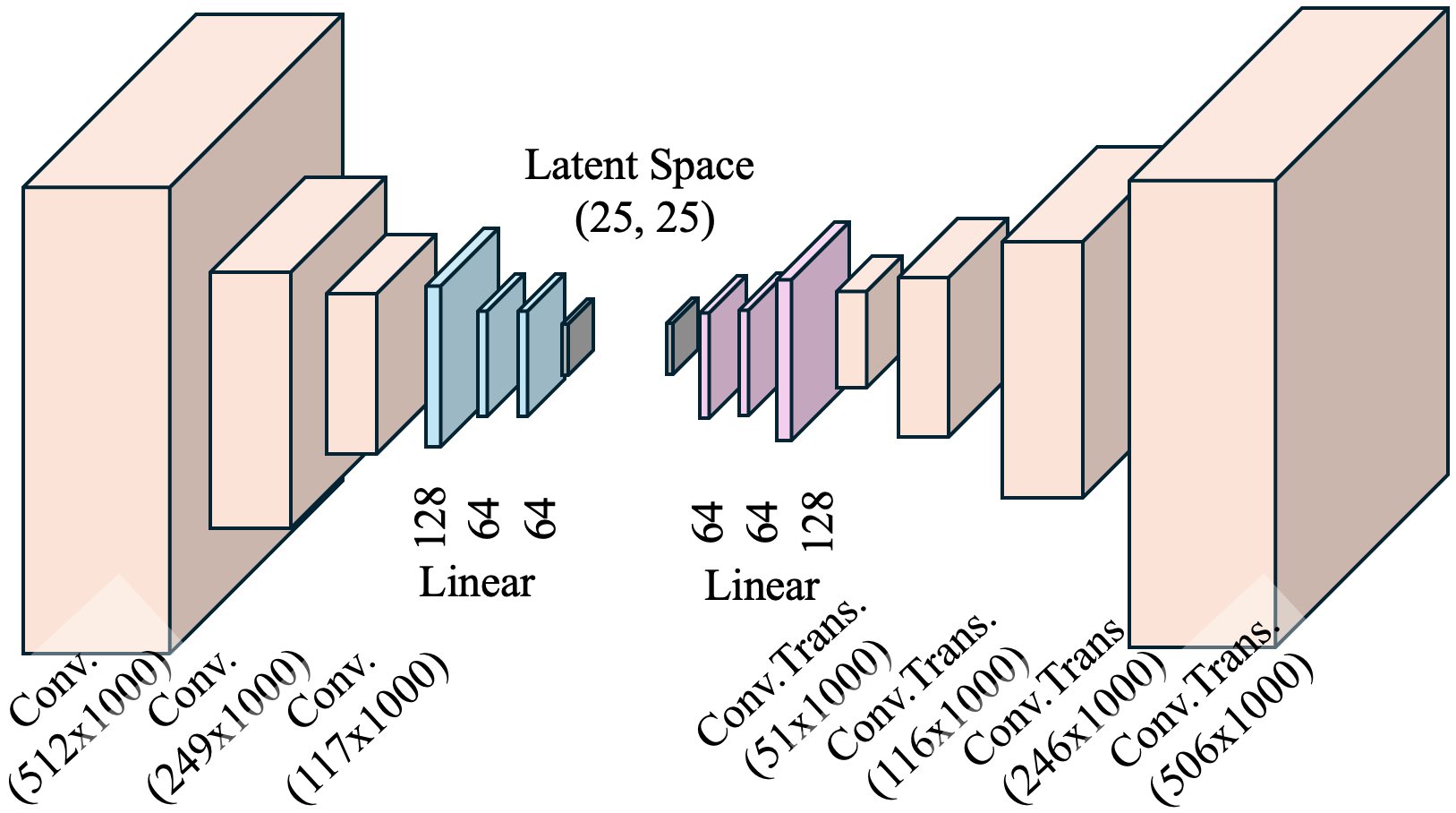}
    \caption{Our CAE comprising 3 hidden and 3 convolutional layers; stride length = 2, except last layer = 1.}
    \label{fig:AEArch:IQConv}
    \vspace{-0.2cm}
\end{figure}

\vspace{+0.1cm}
\textbf{Raw IQ Data.}
\label{label_raw_iq_data}
For IQ data, an input window of either 1\,\textit{ms} or 0.1\,\textit{ms} is employed, resulting in input layer sizes of 5,120 or 512, respectively. These dimensions are manageable for the Edge TPU on the DevBoard Mini. This configuration requires the AE to operate at a rate of $\frac{t}{s}$ per second. However, this is impractical since performance benchmarks for the Edge TPU~\cite{DevBoardMiniBenchmark} indicate that the fastest inference time is approximately 2\,\textit{ms}, even under optimal conditions for models with similar complexity to the AEs. Therefore, in the best-case scenario, the inference time of the Edge TPU is twice the required threshold. The IQ data recording comprises three different scenarios. Initially, the AEs are trained using data from the first two scenarios, which contain simpler data and are therefore easier for the models to learn. Once satisfactory levels of reconstruction and classification accuracy are achieved, the same process is repeated using data from all three scenarios combined. Furthermore, we design two different architectures for evaluation (fully connected and convolutional AE): (1) The first type of AE explored is the \textit{fully connected AE}, where both the encoder and decoder consist exclusively of fully connected layers. All feasible combinations of the number of hidden layers, the number of neurons per hidden layer, and the number of latent variables were generated, resulting in 16 distinct configurations. In cases where the encoder includes three hidden layers, the complete AE architecture comprises ten layers: the encoder consists of an input layer, three hidden layers, and an output layer, totaling five layers. The decoder is a mirror image of the encoder, contributing an additional five layers, making ten layers in total for the AE. Each of the 16 architectures was initially trained for 40 epochs. The five architectures with the lowest reconstruction error were then selected for further training, with each undergoing max.~700 additional epochs to refine performance. (2) As a second type of AE, \textit{convolutional AEs} were implemented. In this architecture, both the encoder and decoder consist of a combination of fully connected layers and convolutional layers. To identify the optimal architecture, or one as close as possible to the best, with respect to reconstruction error, models were trained with all reasonable configurations for 40 epochs. This resulted in 68 distinct combinations. The architecture of the encoder is structured as follows: the convolutional layers are connected sequentially, with the first convolutional layer serving as the input layer of the encoder. The hidden layers are connected in succession, starting with the first hidden layer, followed by the second, and, if applicable, a third hidden layer. For an example of this architecture, see Figure~\ref{fig:AEArch:IQConv}.

\vspace{+0.1cm}
\textbf{Temporal Domain Data.}
\label{label_temporal_data}
For the temporal domain, we utilized only the fully connected AE. For each combination, we trained models with all reasonable architectures for 40 epochs. This involved generating all possible combinations of the number of hidden layers, the number of neurons in each hidden layer, and the various dimensions of the latent variables. The number of hidden layers was constrained to be between 2 and 3, with the number of neurons in each layer being limited to 32, 64, or 128. Additionally, a deeper layer could not have more neurons than a shallower one; for instance, if the first hidden layer had 64 neurons, the second hidden layer could have 32 or 64 neurons, but not 128. This process resulted in 636 distinct combinations. The top 14 models were then selected for further analysis.

\vspace{+0.1cm}
\textbf{Mixed Spectral and Temporal Domain.}
\label{label_spectral_temporal_data}
For the mixed spectral and temporal domain, we employed only the fully connected AE. Initially, the optimal combinations of the recorded values were selected by choosing various combinations of temporal domain data from the top 14 models. For each combination of values identified in the previous step, we construct an AE with the architecture as the best 14 AEs.

\subsection{Compression with Disentanglement}
\label{label_disentanglement}

\begin{figure}[!t]
    \centering
    \includegraphics[width=1.0\linewidth]{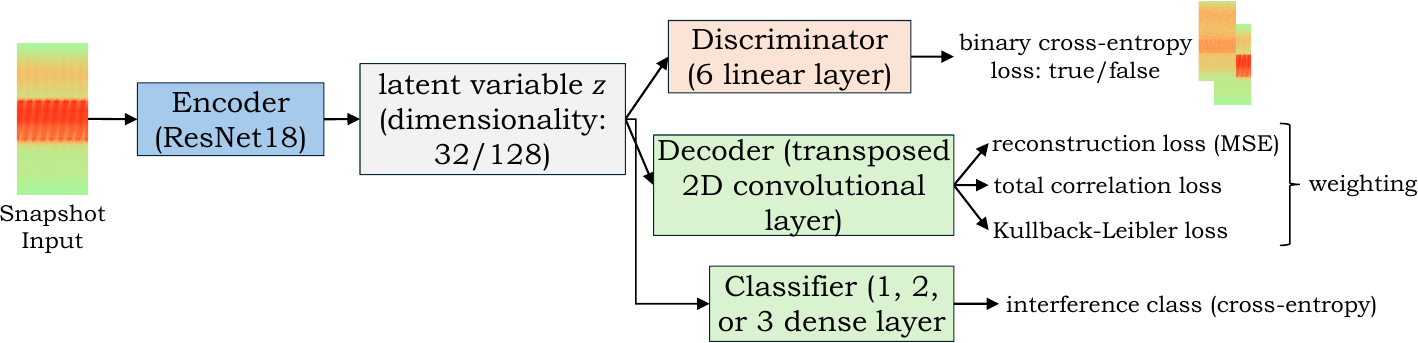}
    \caption{Our VAE that reconstructs and classifies GNSS data.}
    \label{figure_factorvae}
    \vspace{-0.25cm}
\end{figure}

Furthermore, we propose a VAE architecture, based on FactorVAE~\cite{kim_mnih}, to extract significant features from GNSS snapshots in a lower-dimensional space and employ these features for the classification task. To achieve this, we leverage disentanglement, which involves the separation of complex data into distinct, independent factors or components, with each factor representing a specific underlying structure or feature. This technique is widely used in ML to enhance interpretability, generalization, and the extraction of meaningful data representations. The primary objective is to generate new images and train on these images for classification. Our goal is to strike an optimal balance between disentanglement and reconstruction, aiming to improve disentanglement without compromising reconstruction quality. A disentangled representation ensures that each $z_i$ corresponds to exactly one underlying factor. Given that these factors are assumed to vary independently, we aim for a factorial distribution~\cite{kim_mnih}.

\begin{figure}[!t]
    \centering
    \includegraphics[trim=12 12 12 65, clip, width=0.7\linewidth]{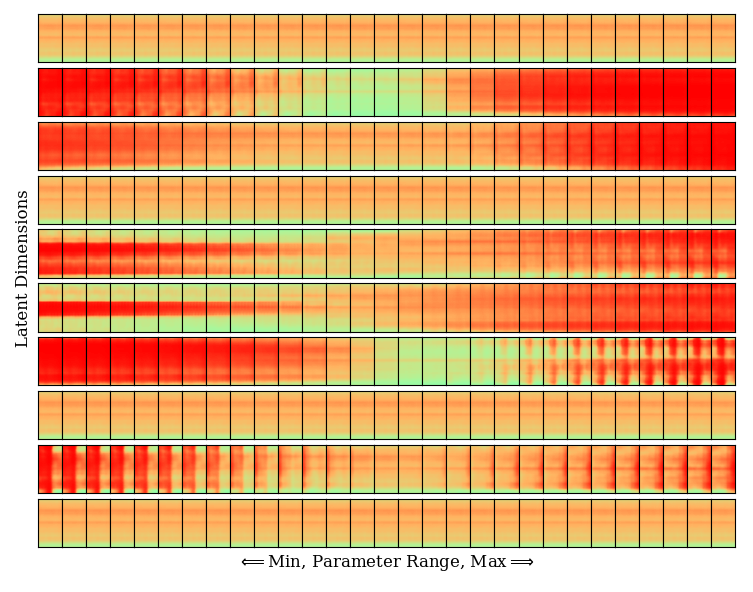}
    \caption{Interpolation of intermediate features for GNSS highway dataset 1. The leftmost and rightmost columns depict the latent features of real-world interference snapshots, while the intermediate snapshots demonstrate interpolations between these images, obtained by varying specific parameters.}
    \label{figure_disentanglement}
    \vspace{-0.25cm}
\end{figure}

Figure~\ref{figure_factorvae} provides an overview of our pipeline. The input consists of a GNSS snapshot with dimensions of $1,024 \times 32$, $1,024 \times 256$, or $512 \times 256$. This is followed by an encoder architecture (i.e., ResNet18) that extracts key features, $z \in \mathbb{R}^d$, with latent dimensionality $d$. We evaluate for $d = 32$ and $d = 128$. Based on the latent variable, a discriminator is trained to classify true or generated snapshots using cross-entropy loss. The decoder is trained to reconstruct the snapshot, employing the mean squared error as the reconstruction loss. The total loss is the sum of the reconstruction loss and the Kullback-Leibler divergence loss. We utilize the optimizer
\begin{equation}
\begin{aligned}
    optim.Adam(&D.parameters(\cdot), lr=lr_D,\\&betas=(beta1_D, beta2_D)),
\end{aligned}
\end{equation}
for the discriminator with parameters $lr_D = 10^{-5}$, $beta1_D = 0.5$, and $beta2_D = 0.9$. We utilize the optimizer
\begin{equation}
\begin{aligned}
    optim.Adam(&VAE.parameters(\cdot), lr=lr_{VAE},\\&betas=(beta1_{VAE}, beta2_{VAE}))
\end{aligned}
\end{equation}
for the VAE encoder and decoder with parameters $lr_{VAE} = 10^{-4}$, $beta1_{VAE} = 0.9$, and $beta2_{VAE} = 0.999$. The model is trained for 250 epochs. Finally, the latent variables are extracted, and a network (i.e., a series of linear layers $[1, 2, 3]$) is trained, incorporating batch normalization or ReLU activation functions between layers. We conduct an extensive hyperparameter search across all parameters, see Figure~\ref{figure_disentanglement}, for an exemplary interpolation inbetween latent features.

\section{Experiments}
\label{sec:experiments}

\begin{figure*}[!t]
    \centering
	\begin{minipage}[t]{0.325\linewidth}
        \centering
    	\includegraphics[width=1.0\linewidth]{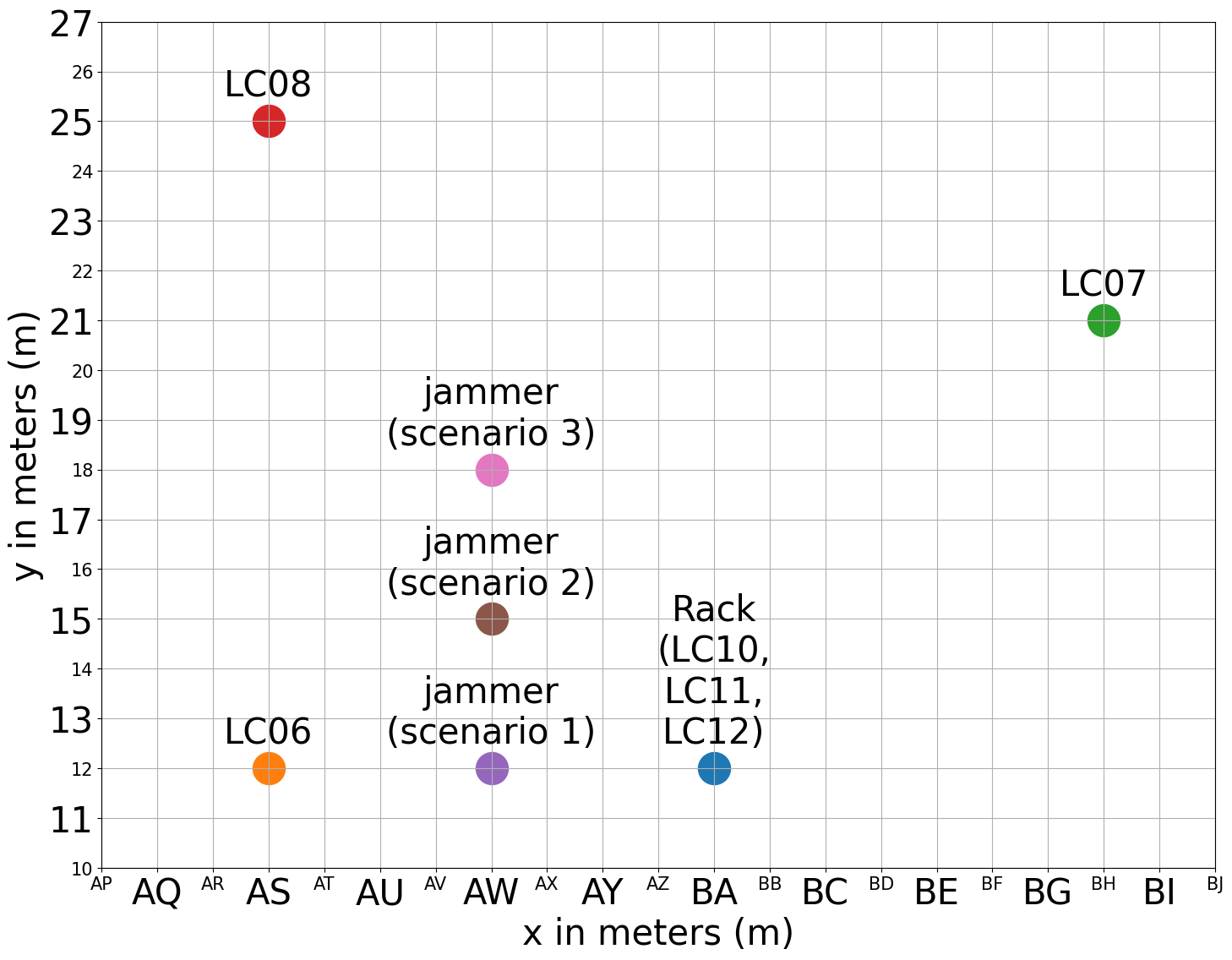}
        \subcaption{Setup for the first recording.}
        \label{fig:Time_Freq_Setup_day_1}
    \end{minipage}
    \hfill
	\begin{minipage}[t]{0.325\linewidth}
        \centering
    	\includegraphics[width=1.0\linewidth]{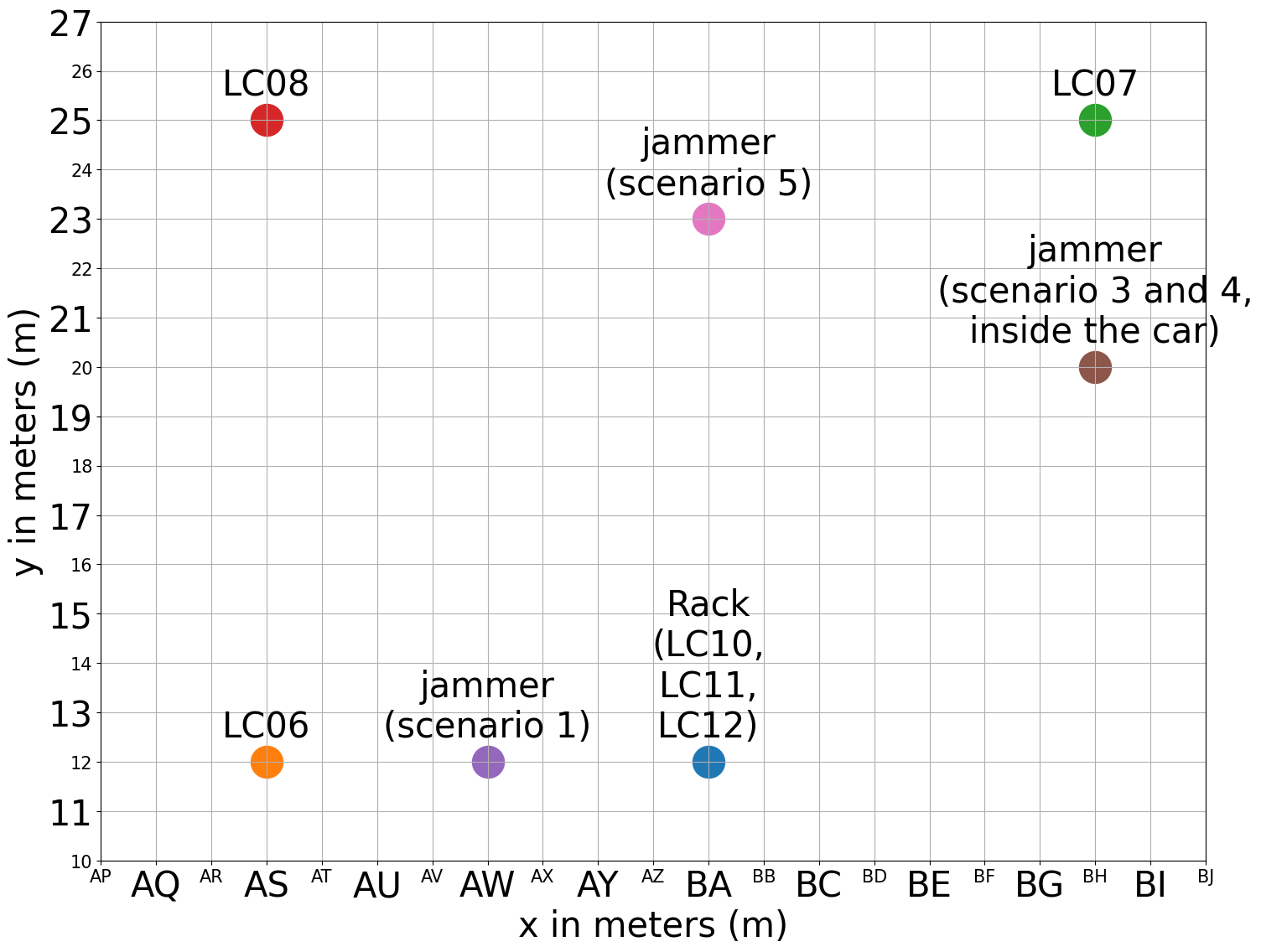}
        \subcaption{Setup for the second recording.}
        \label{fig:Time_Freq_Setup_day_2}
    \end{minipage}
    \hfill
	\begin{minipage}[t]{0.325\linewidth}
        \centering
    	\includegraphics[width=1.0\linewidth]{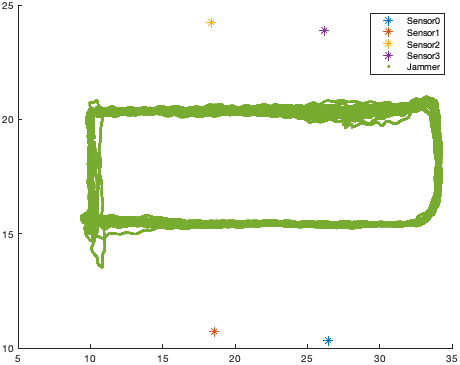}
        \subcaption{Trajectory for the second recording.}
        \label{fig:Tima_Freq_Setup_day_dynamic_2}
    \end{minipage}
    \caption{Data acquisition setups for the spectral and temporal domain.}
    \label{fig:scenarios_Time_Freq}
    \vspace{-0.25cm}
\end{figure*}

\begin{figure}[!bp]
    \centering
	\begin{minipage}[t]{0.24\linewidth}
        \centering
    	\includegraphics[width=1.0\linewidth]{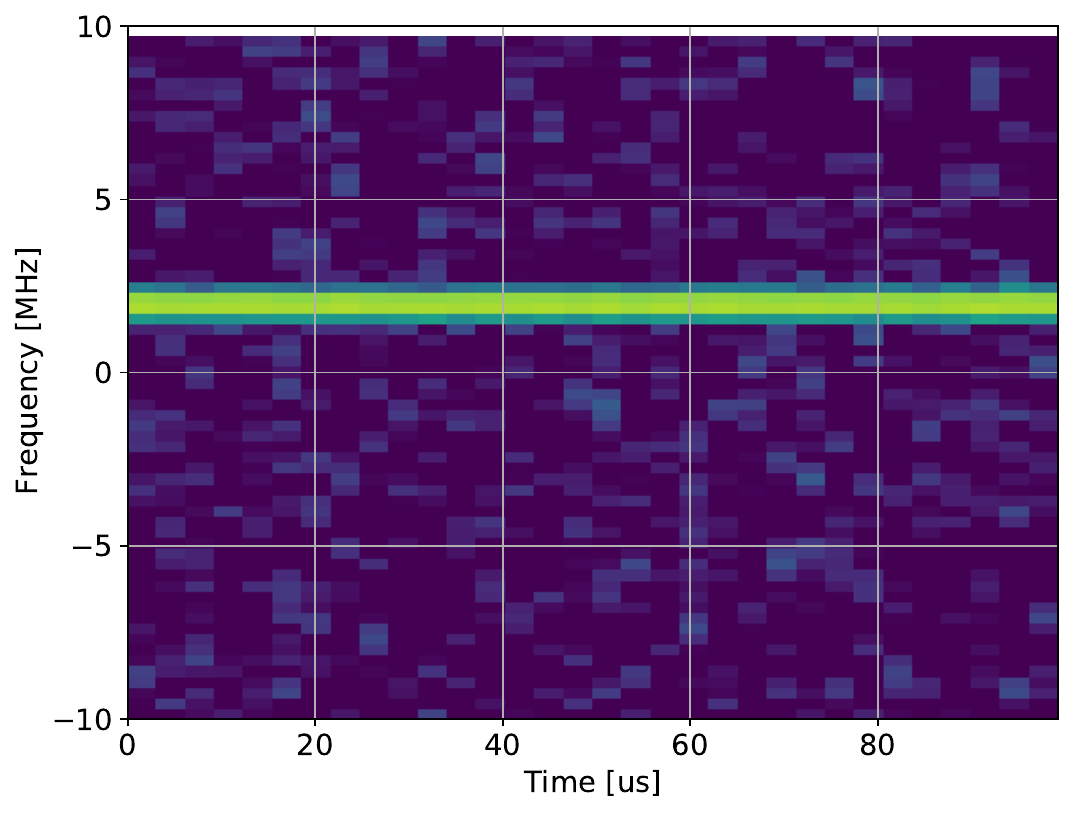}
        \subcaption{Singletone.}
        \label{fig:classexample:cw}
    \end{minipage}
    \hfill
	\begin{minipage}[t]{0.24\linewidth}
        \centering
    	\includegraphics[width=1.0\linewidth]{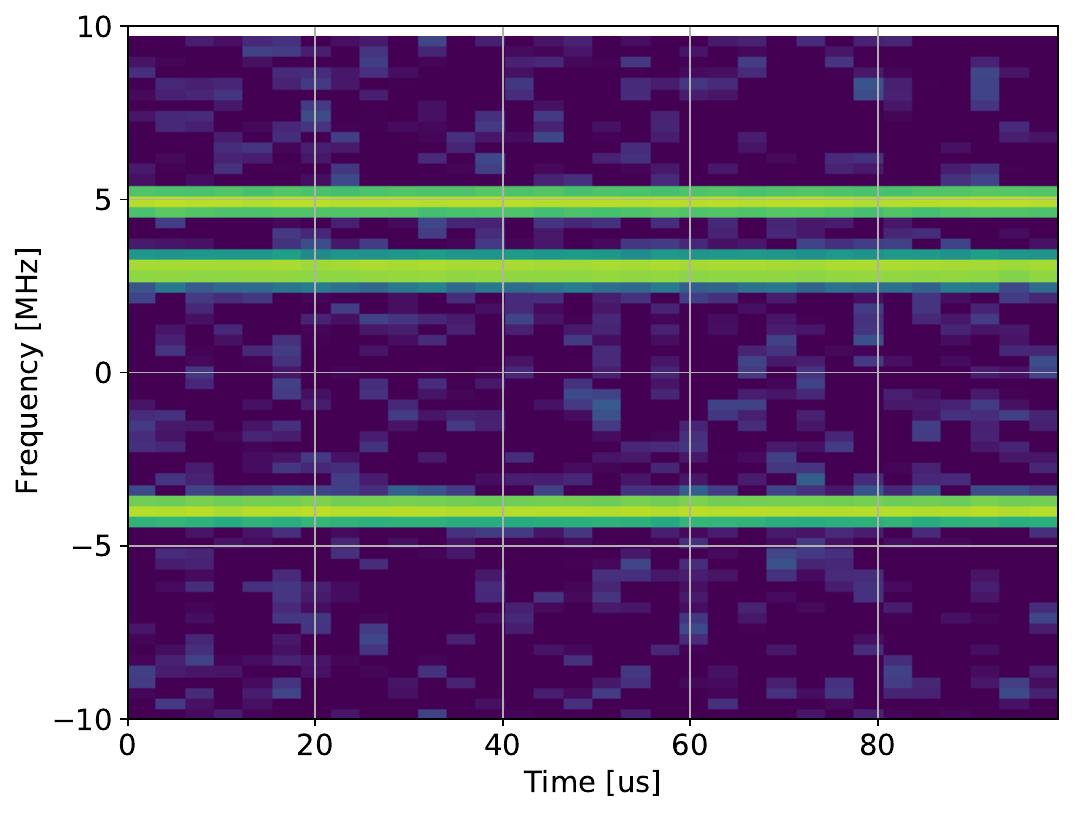}
        \subcaption{Multitone.}
        \label{fig:classexample:multi}
    \end{minipage}
    \hfill
	\begin{minipage}[t]{0.24\linewidth}
        \centering
    	\includegraphics[width=1.0\linewidth]{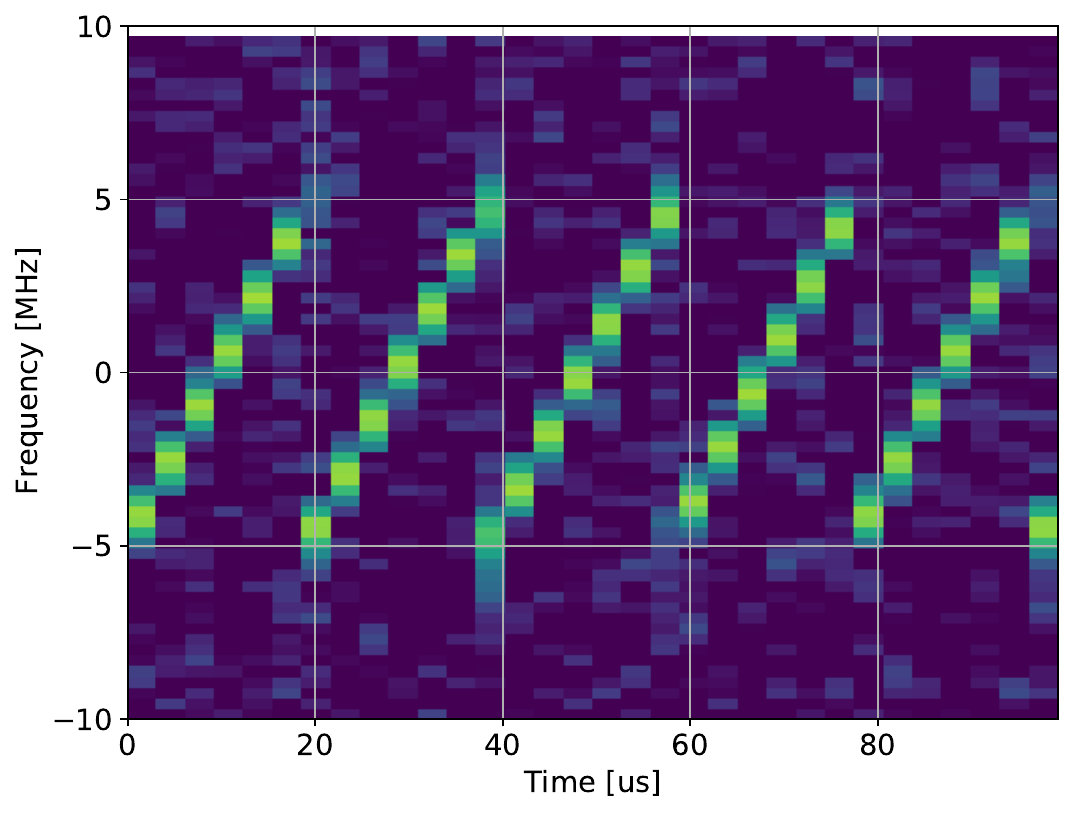}
        \subcaption{Chirp.}
        \label{fig:classexample:chirp}
    \end{minipage}
    \hfill
	\begin{minipage}[t]{0.24\linewidth}
        \centering
    	\includegraphics[width=1.0\linewidth]{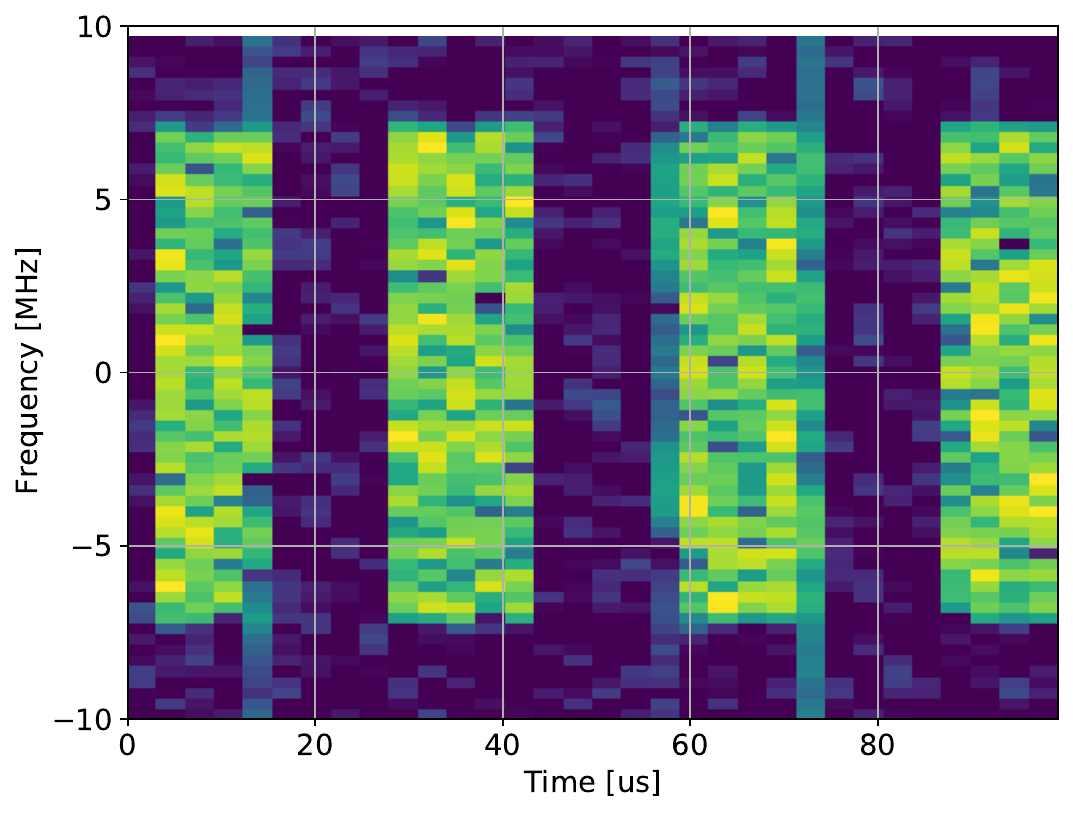}
        \subcaption{Pulsed noise.}
        \label{fig:classexample:pulsed}
    \end{minipage}
    \caption{Example spectrograms of different waveform types~\cite{Merwe2022}.}
    \label{fig:interferences}
\end{figure}

\subsection{Datasets}

The data acquisition process encompassed the collection of three types of data: spectral, temporal, and raw IQ data. Multiple datasets were recorded and subsequently utilized for experimental purposes.

(1) The data collection was conducted at the Fraunhofer IIS L.I.N.K. test and application center in Nürnberg, Germany, which offers an interference-free environment suitable for the secure use of GNSS jammers. Spectral and temporal data were recorded over two days, encompassing various (non-)multipath scenarios using six strategically positioned low-cost GNSS sensors, as described by Brieger et al.~\cite{Brieger2022Classification}. Figure~\ref{fig:scenarios_Time_Freq} illustrates the experimental setup. On the first day, scenarios were configured to minimize multipath effects by removing barriers between the jammer and sensors. On the second day, movable walls were introduced to replicate real-world signal reflections. As depicted in Figure~\ref{fig:Time_Freq_Setup_day_1}, six LC sensors were deployed, with three mounted on a rack. The jammer was placed at multiple positions, altering its distance from each sensor (see Figure~\ref{fig:Time_Freq_Setup_day_1}). For each jammer position, 33 distinct types of interference signals were recorded, categorized into seven waveform classes: Noise, Chirp, Multitone, Pulsed, Frequency Hopper, and Modulated, as shown in Figure~\ref{fig:interferences}. Regarding the raw IQ data, three scenarios were captured with the jammer positioned 22\,\textit{m} from the sensor. The dataset comprises 5-minute recordings for each scenario, including one with a clean signal and two with jammers operating at attenuation levels of -20\,dB and -26\,dB, respectively, offering varying interference intensities.

(2) Highway dataset 1~\cite{heublein_raichur_ion,ott_heublein_icl}: a sensor station was strategically positioned on a bridge along a highway to capture short, wideband snapshots within the E1 and E6 GNSS frequency bands. The system recorded 20\,\textit{ms} raw IQ snapshots triggered by energy detection, operating at a sample rate of 62.5\,MHz, an analog bandwidth of 50\,MHz, and an 8-bit resolution. The resulting spectrogram images have dimensions of $512 \times 243$. Data streams were manually analyzed by experts using thresholding techniques applied to C/N0 and AGC values, leading to the classification of snapshots into 11 distinct classes.

(3) Additionally, an extensive GNSS dataset comprising 72 different interference types was recorded~\cite{heublein_feigl_crpa}. The experimental setup positioned a receiver antenna at one end of the hall and an MXG vector signal generator at the opposite end. Data acquisition was conducted under a variety of configurations, including scenarios within an unoccupied environment and setups involving absorber walls placed between the antenna and the generator. This arrangement enabled the capture of distinct multipath effects, ranging from minimal to pronounced reflections, as well as cases of significant signal absorption. Further details are available in Heublein et al.~\cite{heublein_feigl_crpa}.
%\vspace{-0.1cm}

\subsection{Hardware Setup}

For the inference of the compiled TensorFlow Lite model, the DevBoard Mini~\cite{DevBoardMini} was utilized, featuring a MediaTek 8167s SoC (Quad-core Arm Cortex-A35), 2\,GB of RAM, and a surface-mounted module incorporating the Edge TPU. This configuration enables real-time inference, facilitating efficient processing and classification of compressed GNSS signals. Additionally, Nvidia Tesla V100-SXM2 GPUs with 32 GB of VRAM were employed to train the large-scale (V)AE models, significantly accelerating the model development process.
\section{Results}
\label{sec:results}

\subsection{Detection \& classification Accuracy}
\label{detandclasacc}

A trade-off inherently exists between the accuracy of GNSS interference detection and classification and the compressibility of models. This study emphasizes model compressibility to reduce the power consumption of the Edge TPU and minimize the bandwidth required for data transmission. The subsequent sections present the performance of the MIAS framework across different domains: the spectral domain (Section~\ref{label_mias_spectral}), the temporal domain (Section~\ref{label_mias_temporal}), and the combined spectral and temporal domains (Section~\ref{label_mias_combination}).\vspace{+0.25cm}

\textbf{MIAS Performance on Spectral Domain.}
\label{label_mias_spectral}
\begin{figure}[!bp]
    \centering
	\begin{minipage}[t]{0.49\linewidth}
        \centering
    	\includegraphics[trim=0 0 0 78, clip, width=1\linewidth]{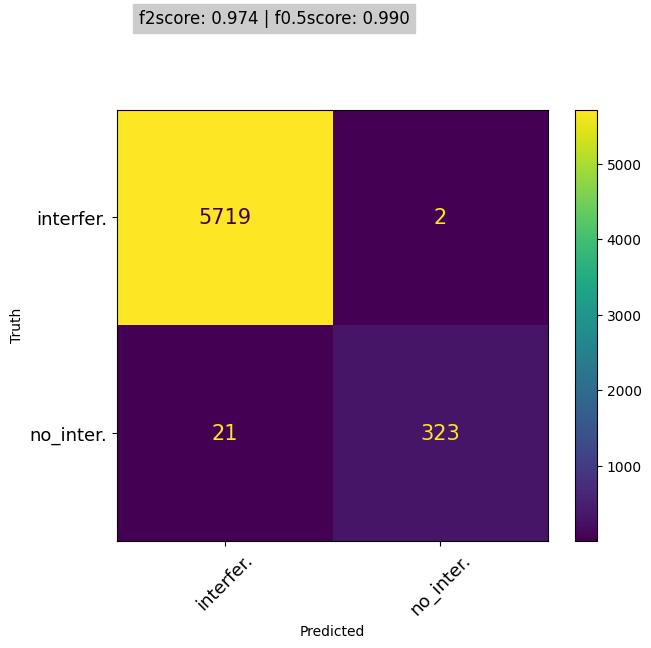}
        \subcaption{Raw data ($\text{F}_{2}$-score: 0.974, $\text{F}_{0.5}$-score: 0.990).}
    \end{minipage}
    \hfill
	\begin{minipage}[t]{0.49\linewidth}
        \centering
    	\includegraphics[trim=0 0 0 78, clip, width=1\linewidth]{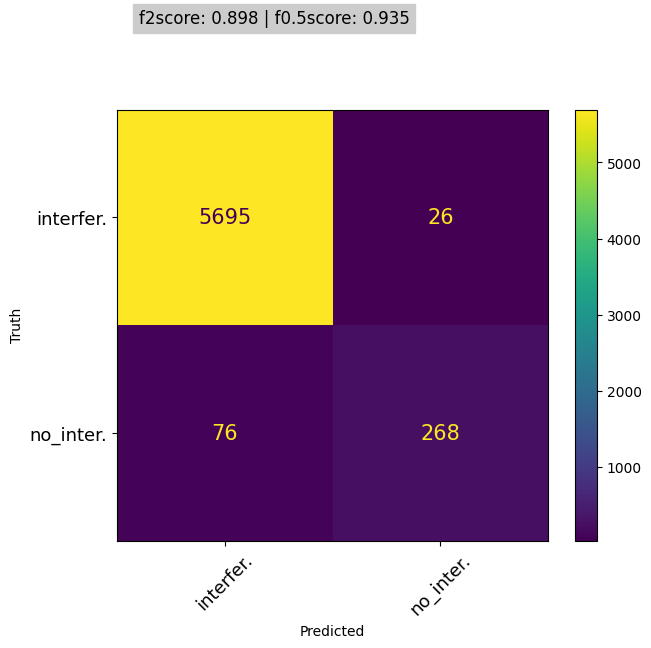}
        \subcaption{Reconstructed data ($\text{F}_{2}$-score: 0.898, $\text{F}_{0.5}$-score: 0.935).}
    \end{minipage}
    \caption{Detection accuracy on spectral domain.}
    \label{fig:sdr_per_best_cms_det}
\end{figure}
\begin{figure*}[!t]
    \centering
	\begin{minipage}[t]{0.235\linewidth}
        \centering
    	\includegraphics[width=1\linewidth]{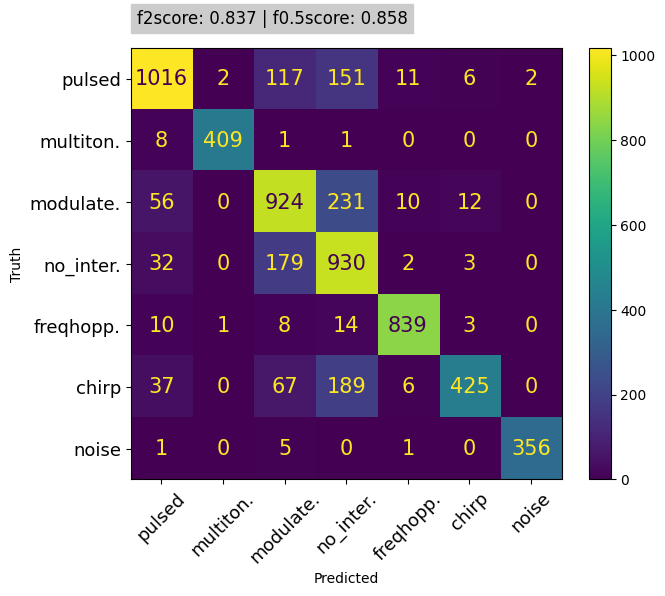}
        \subcaption{Raw data small.}
    \end{minipage}
    \hfill
	\begin{minipage}[t]{0.235\linewidth}
        \centering
    	\includegraphics[width=1\linewidth]{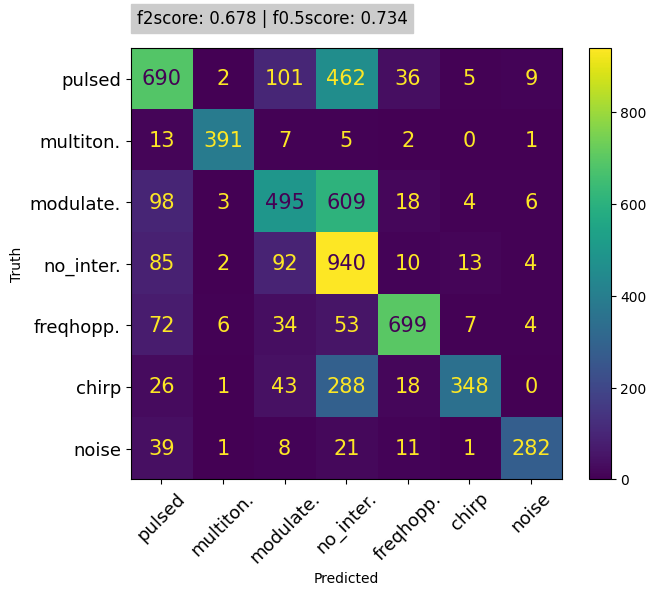}
        \subcaption{Reconstructed data small.}
    \end{minipage}
    \hfill
	\begin{minipage}[t]{0.235\linewidth}
        \centering
    	\includegraphics[width=1\linewidth]{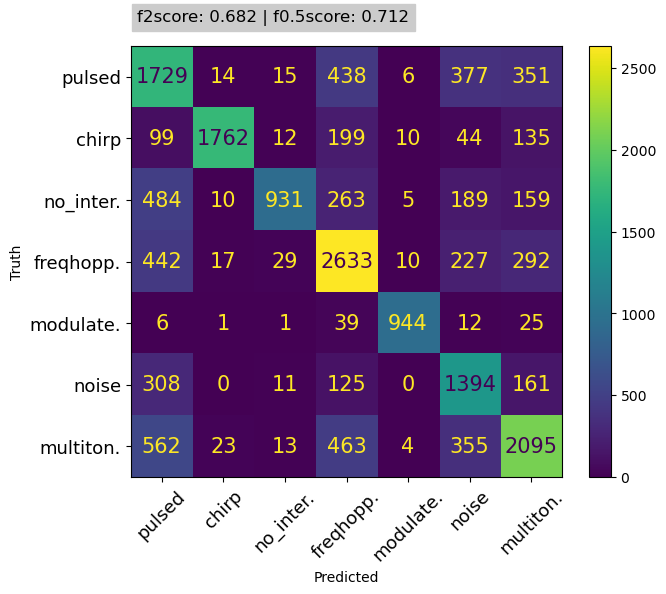}
        \subcaption{Raw data full.}
    \end{minipage}
    \hfill
	\begin{minipage}[t]{0.235\linewidth}
        \centering
    	\includegraphics[width=1\linewidth]{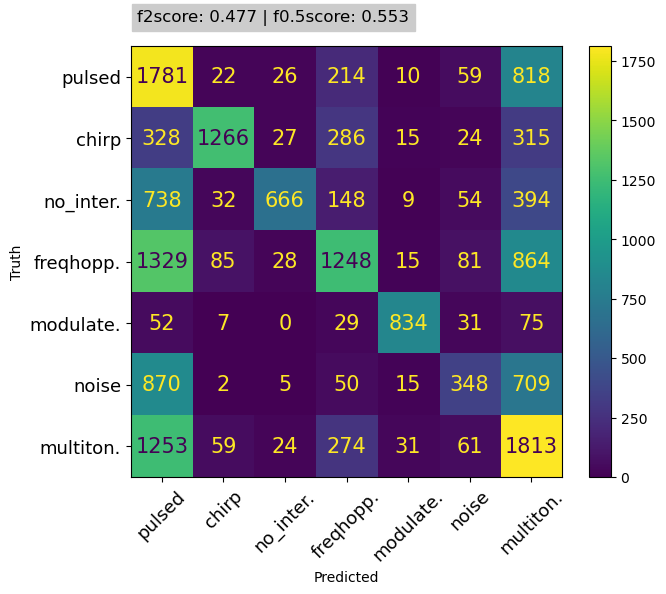}
        \subcaption{Reconstructed data full.}
    \end{minipage}
    \caption{Classification accuracy on spectral domain.}
    \label{fig:sdr_per_best_cms_clas}
\end{figure*}
\begin{figure*}[!t]
    \centering
	\begin{minipage}[t]{0.235\linewidth}
        \centering
    	\includegraphics[width=1\linewidth]{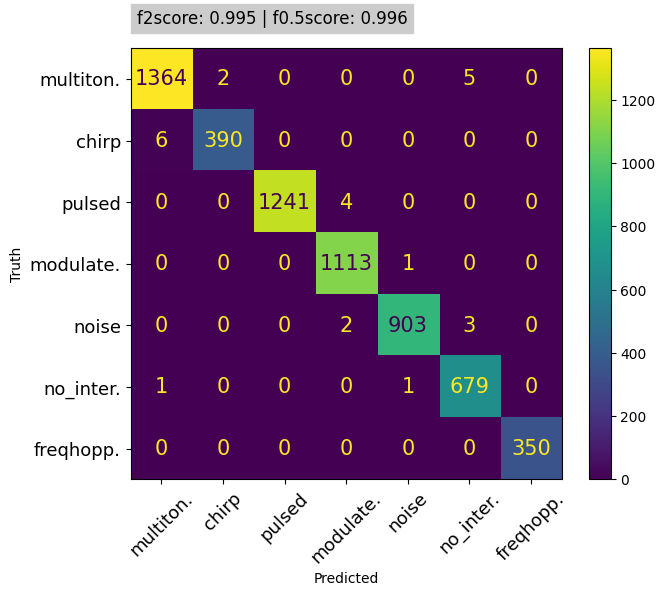}
        \subcaption{Raw data small.}
    \end{minipage}
    \hfill
	\begin{minipage}[t]{0.235\linewidth}
        \centering
    	\includegraphics[width=1\linewidth]{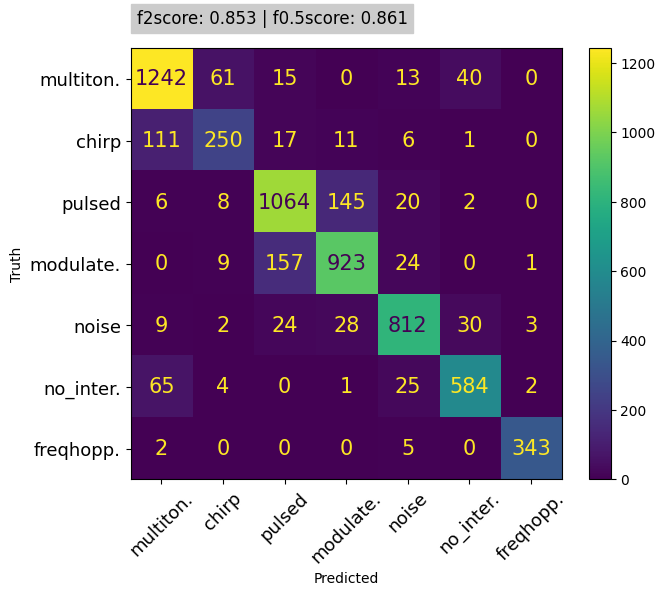}
        \subcaption{Reconstructed data small.}
    \end{minipage}
    \hfill
	\begin{minipage}[t]{0.235\linewidth}
        \centering
    	\includegraphics[width=1\linewidth]{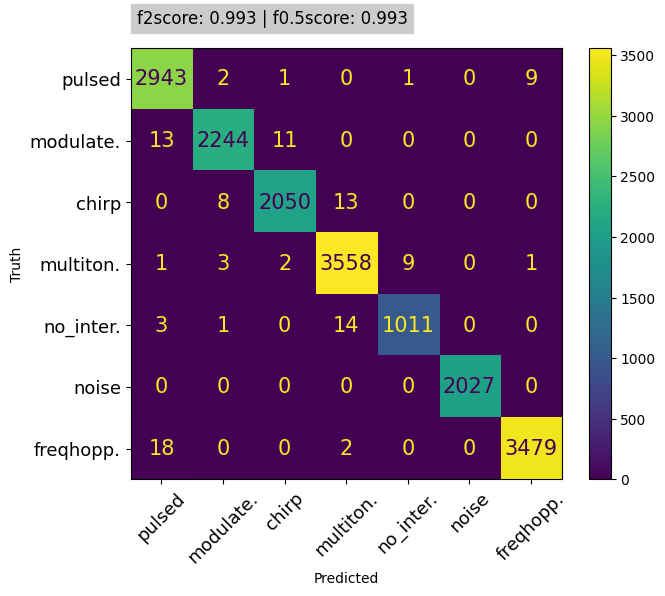}
        \subcaption{Raw data full.}
    \end{minipage}
    \hfill
	\begin{minipage}[t]{0.235\linewidth}
        \centering
    	\includegraphics[width=1\linewidth]{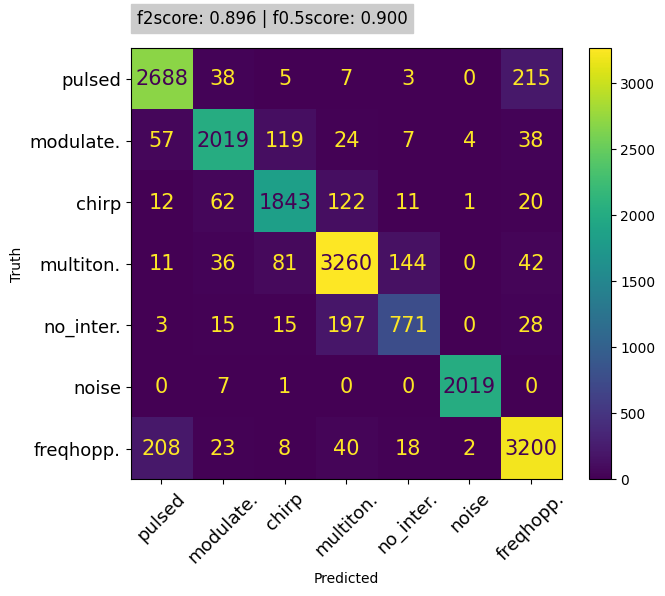}
        \subcaption{Reconstructed data full.}
    \end{minipage}
    \caption{Confusion matrices for the classification for the temporal domain.}
    \label{fig:time_per_best_cms_classification}
    \vspace{-0.25cm}
\end{figure*}
The detection and classification performance of the MIAS framework in the spectral domain is evaluated using two datasets: a small dataset and a full dataset. Figure~\ref{fig:sdr_per_best_cms_det} illustrates the detection accuracy for the small dataset, where the detection pipeline achieves an $\text{F}_{2}$-score of 0.974 for the original data and 0.898 for the reconstructed data, indicating an accuracy loss of approximately 0.076 $\text{F}_{2}$. For classification, the $\text{F}_{2}$-score for the original data is 0.837 (Figure~\ref{fig:sdr_per_best_cms_clas}), decreasing to 0.678 due to accuracy degradation caused by the reconstruction of the quantized AE. On the full dataset, the classification pipeline achieves an $\text{F}_{2}$-score of 0.682 for raw data, which further declines to 0.477 for reconstructed data. These findings indicate that spectral domain data alone lacks sufficient information for a simple detection and classification pipeline. However, integrating spectral data into the classification process with other domain data may still enhance overall information availability.

\newpage\textbf{MIAS Performance on Temporal Domain.}
\label{label_mias_temporal}
\begin{figure}[!bp]
    \centering
	\begin{minipage}[t]{0.49\linewidth}
        \centering
    	\includegraphics[trim=0 0 0 78, clip, width=1\linewidth]{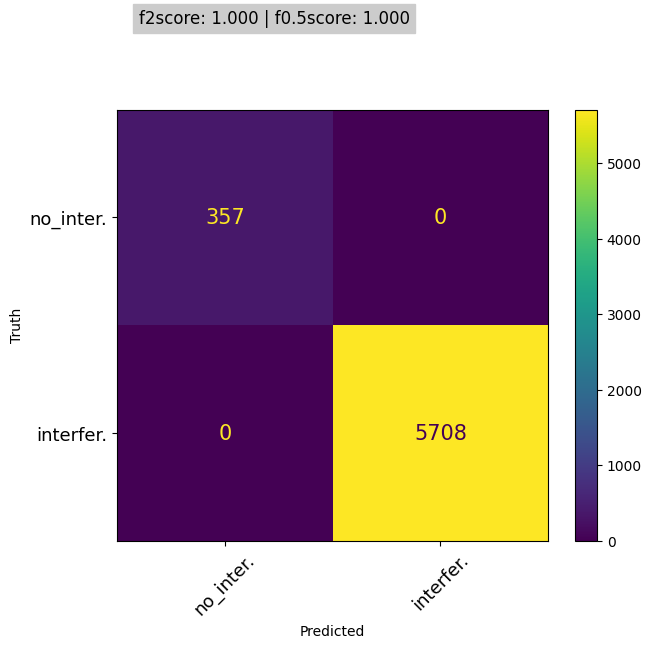}
        \subcaption{Raw data small (F-score: 1.0, $\text{F}_{0.5}$-score: 1.0).}
    \end{minipage}
    \hfill
	\begin{minipage}[t]{0.49\linewidth}
        \centering
    	\includegraphics[trim=0 0 0 78, clip, width=1\linewidth]{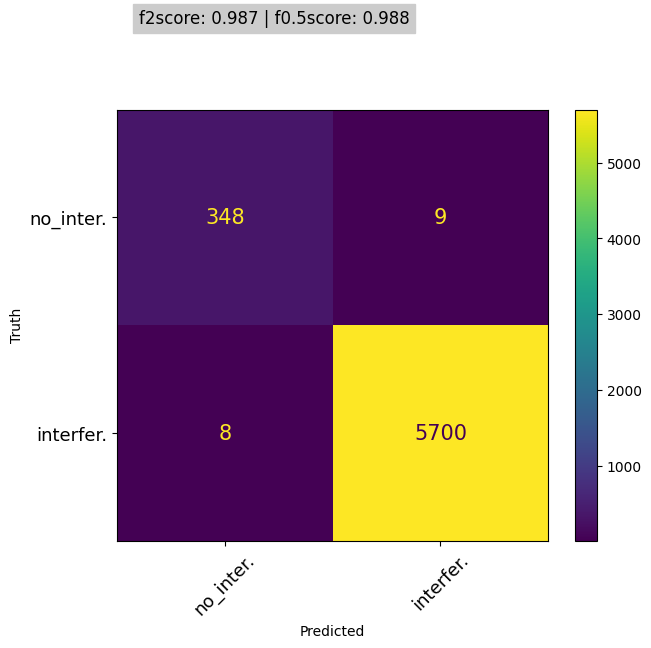}
        \subcaption{Reconstructed data small (F-score: 0.987, $\text{F}_{0.5}$-score: 0.988).}
    \end{minipage}
    \caption{Detection accuracy on temporal domain.}
    %\vspace{-0.5cm}
\label{fig:time_per_best_cms_detection}
%\vspace{-0.5cm}
\end{figure}
The detection accuracy for the small dataset achieves an $\text{F}_{2}$-score of 0.1 for raw data and 0.987 for reconstructed data (Figure~\ref{fig:time_per_best_cms_detection}), indicating that the AE captures the internal structure of the temporal domain more effectively than the spectral domain. This suggests that temporal data contains more usable information. High accuracy on the full dataset further supports the conclusion that temporal data provides sufficient information for interference detection, even with complex scenarios involving varying jammer distances and multipath effects. The classification pipeline achieves an $\text{F}_{2}$-score of 0.995 (Figure~\ref{fig:time_per_best_cms_classification}), although accuracy decreases for reconstructed data due to quantization losses in the AE.
%\vspace{+0.5cm}

\vspace{+0.1cm}
\textbf{MIAS Performance on Temporal and Spectral Domains.}
\label{label_mias_combination}
\begin{figure*}[!t]
    \centering
	\begin{minipage}[t]{0.235\linewidth}
        \centering
    	\includegraphics[trim=0 0 0 78, clip, width=1\linewidth]{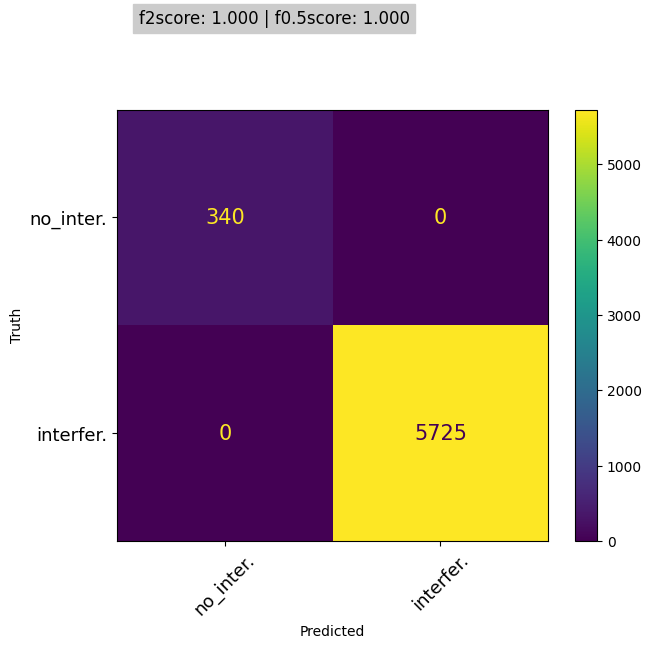}
        \subcaption{Raw data small (F-score: 1.0, $\text{F}_{0.5}$-score: 1.0).}
    \end{minipage}
    \hfill
	\begin{minipage}[t]{0.235\linewidth}
        \centering
    	\includegraphics[trim=0 0 0 78, clip, width=1\linewidth]{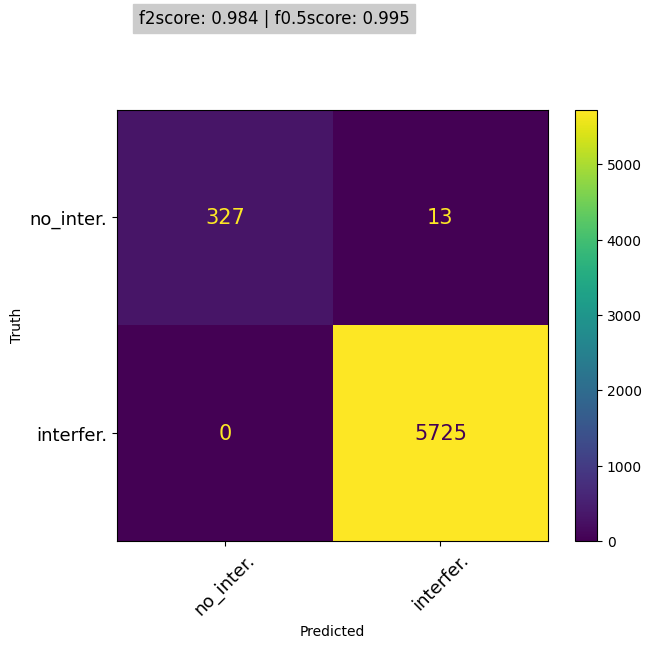}
        \subcaption{Reconstructed data small (F-score: 0.984, $\text{F}_{0.5}$-score: 0.995).}
    \end{minipage}
    \hfill
	\begin{minipage}[t]{0.235\linewidth}
        \centering
    	\includegraphics[trim=0 0 0 78, clip, width=1\linewidth]{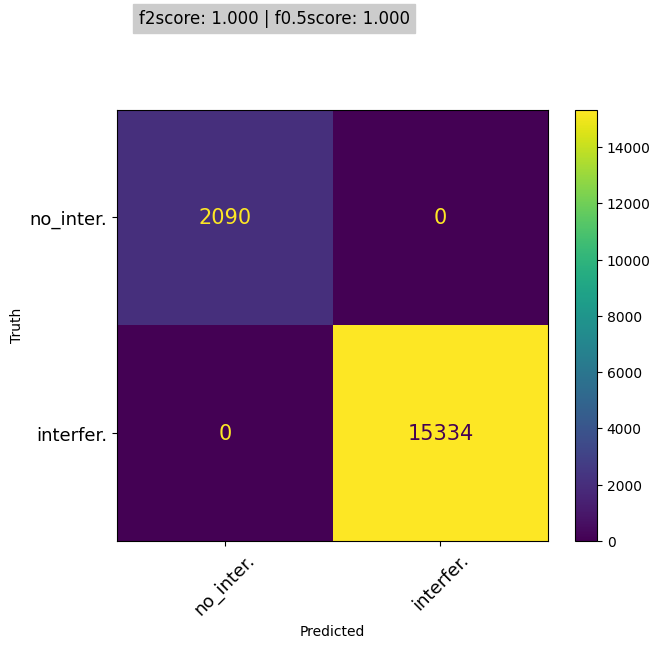}
        \subcaption{Raw data full (F-score: 1.0, $\text{F}_{0.5}$-score: 1.0).}
    \end{minipage}
    \hfill
	\begin{minipage}[t]{0.235\linewidth}
        \centering
    	\includegraphics[trim=0 0 0 78, clip, width=1\linewidth]{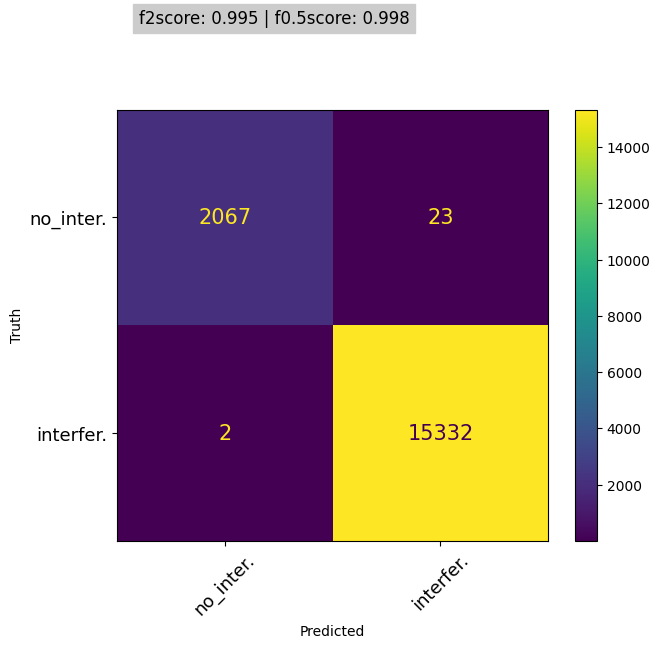}
        \subcaption{Reconstructed data full (F-score: 0.995, $\text{F}_{0.5}$-score: 0.998).}
    \end{minipage}
	\begin{minipage}[t]{0.235\linewidth}
        \centering
    	\includegraphics[width=1\linewidth]{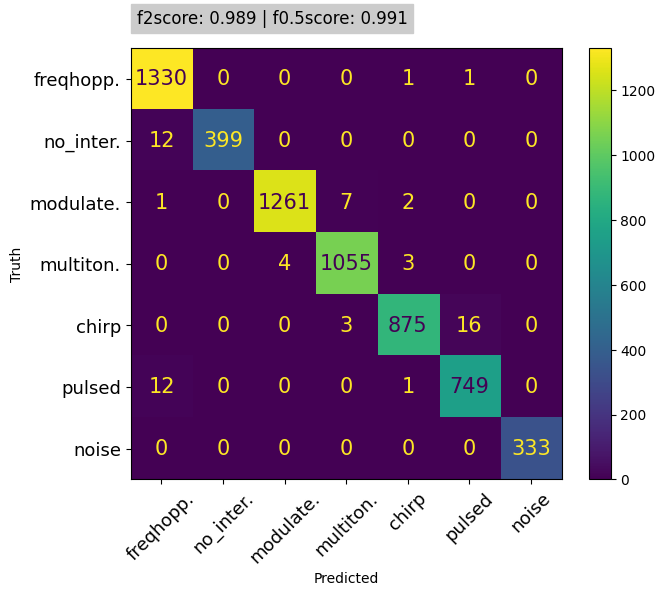}
        \subcaption{Raw data small.}
    \end{minipage}
    \hfill
	\begin{minipage}[t]{0.235\linewidth}
        \centering
    	\includegraphics[width=1\linewidth]{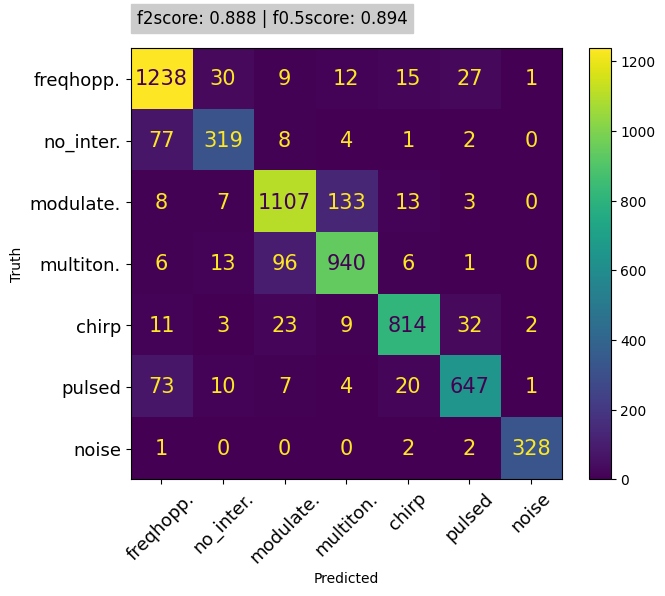}
        \subcaption{Reconstructed data small.}
    \end{minipage}
    \hfill
	\begin{minipage}[t]{0.235\linewidth}
        \centering
    	\includegraphics[width=1\linewidth]{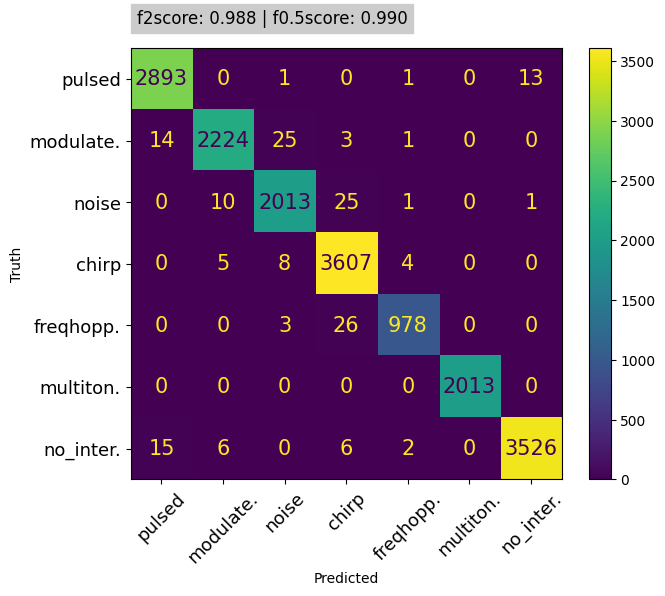}
        \subcaption{Raw data full.}
    \end{minipage}
    \hfill
	\begin{minipage}[t]{0.235\linewidth}
        \centering
    	\includegraphics[width=1\linewidth]{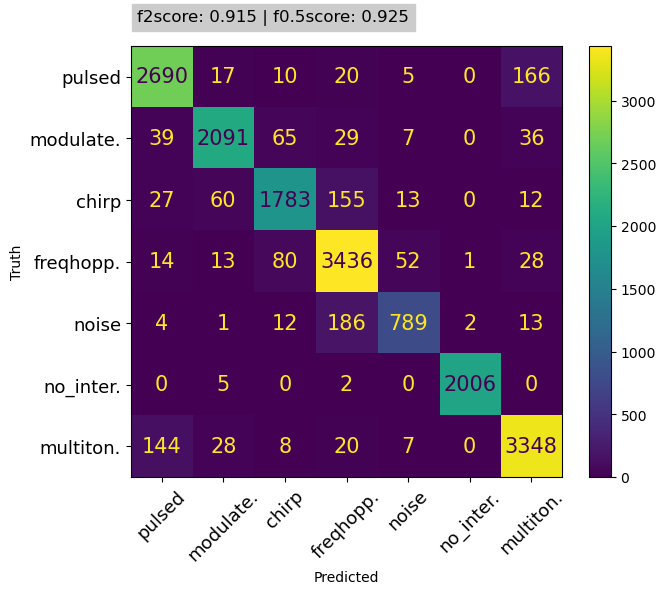}
        \subcaption{Reconstructed data full.}
    \end{minipage}
    \caption{Confusion matrices for the detection (top) and classification (bottom) on the temporal and spectral domain mixed.}
    \label{fig:timesdr_per_best_cms_dection_cl}
\end{figure*}
\begin{figure*}[!t]
    \centering
	\begin{minipage}[t]{1.0\linewidth}
        \centering
    	\includegraphics[width=1.0\linewidth]{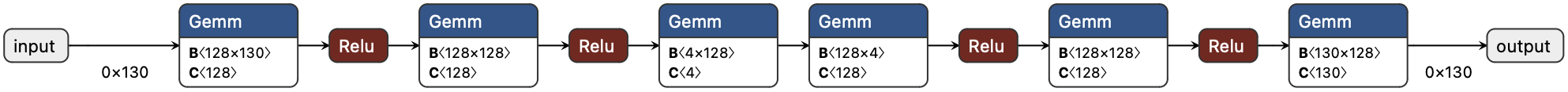}
    \end{minipage}
    \hfill
	\begin{minipage}[t]{1.0\linewidth}
        \centering
    	\includegraphics[width=1.0\linewidth]{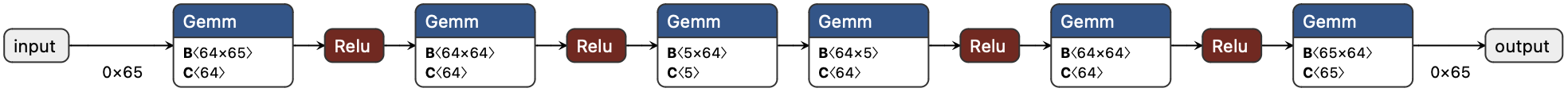}
    \end{minipage}
    \hfill
	\begin{minipage}[t]{1.0\linewidth}
        \centering
    	\includegraphics[width=1.0\linewidth]{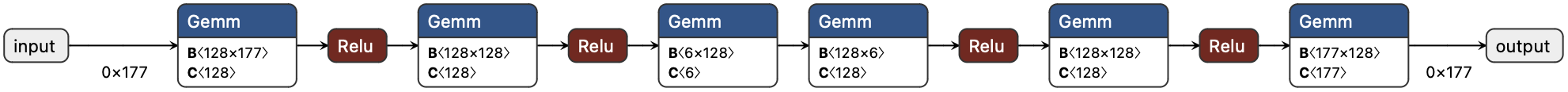}
    \end{minipage}
    \caption{Best AE architectures for the spectral (top), temporal (middle), and spectral and temporal combined (bottom) domains.}
    \label{fig:AE_best_architectures}
\end{figure*}
The temporal domain data alone provides substantial information, leading to strong detection and classification accuracy for the full dataset. When combined with spectral domain data, the detection pipeline achieves an $\text{F}_{2}$-score of 1.0 for both raw and reconstructed data (Figure~\ref{fig:timesdr_per_best_cms_dection_cl}), with no additional accuracy gain from the spectral data. For the full dataset, the detection pipeline also maintains an $\text{F}_{2}$-score of 1.0, and the addition of spectral data does not improve performance. The classification pipeline achieves an $\text{F}_{2}$-score of 0.989 for the small dataset's raw data, and 0.988 for the full dataset's raw data. For reconstructed data, the $\text{F}_{2}$-score drops to 0.915, reflecting a loss of 0.073 due to reconstruction accuracy loss in the quantized AE. Although spectral domain data provides a slight accuracy improvement, the combination of temporal and spectral data enhances the overall information.

\subsection{Energy Saving of MIAS}
\label{MiasEnergy}

The power consumption of the LC COTS system, as introduced in \cite{Merwe2022}, primarily depends on WiFi and networking. To achieve power savings, the data must be compressed to a size where the energy saved in networking exceeds the power consumption of the Edge TPU. Since the power consumption formula is based on a batch size of 1,000, the pipeline requires 1,000 inputs before processing. The TPU consumes 1.6\,W for each batch of 1,000, equating to 1.6\,Ws per 1,000s, or 4.44\,$\mu$Wh. WiFi and networking power consumption is 394 \,mWh~\cite{Merwe2022}, so a 1\% compression would save about 4\,mWh, which is 1,000 times greater than the TPU's power consumption. The system collects 253 data points per second, and the best AE for the combined temporal and spectral domains compresses 177 values to 6, reducing the required data transmission by 67\%. As a result, the total power consumption of WiFi and networking drops to 264\,mWh, saving 130\,mWh, which is 29,000 times the power consumption of the TPU over the same period.

\subsection{Evaluating Power Consumption of AEs}
\label{EvalPowerCon}

\begin{figure*}[!t]
    \centering
	\begin{minipage}[t]{0.325\linewidth}
        \centering
    	\includegraphics[trim=10 10 10 10, clip, width=1.0\linewidth]{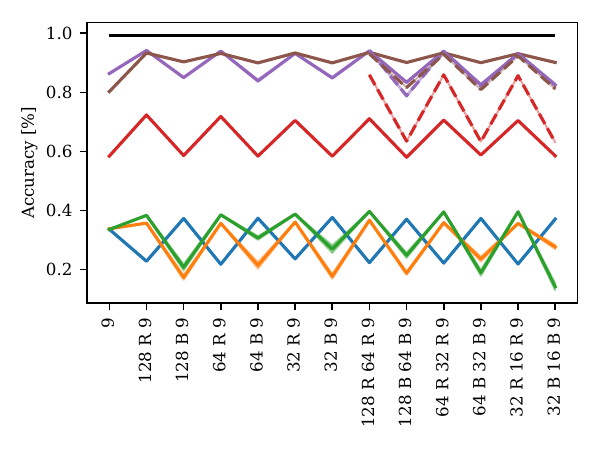}
        \subcaption{Small-scale indoor dataset~\cite{Brieger2022Classification}.}
    \end{minipage}
    \hfill
	\begin{minipage}[t]{0.325\linewidth}
        \centering
    	\includegraphics[trim=10 10 10 10, clip, width=1.0\linewidth]{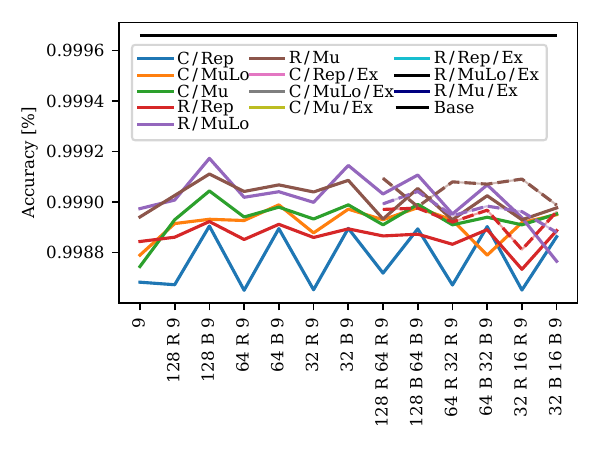}
        \subcaption{Highway dataset 1~\cite{ott_heublein_icl}.}
    \end{minipage}
    \hfill
	\begin{minipage}[t]{0.325\linewidth}
        \centering
    	\includegraphics[trim=10 10 10 10, clip, width=1.0\linewidth]{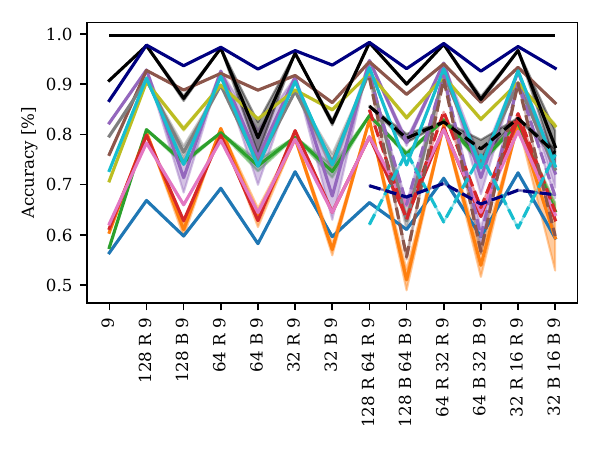}
        \subcaption{Large-scale indoor dataset~\cite{heublein_feigl_crpa}.}
    \end{minipage}
    \caption{Overview of results for the factorized VAE with disentanglement for compression. Number x-ticks: output dimension of the dense layer, B: batch normalization, R: ReLU activation, C: convolutional encoder, R: ResNet18 encoder, Rep: reparapetrization, Mu: $\mu$ average, Lo: $\log$ variance, Ex: extended to quadratic input.}
    \label{label_disentanglement_results}
\end{figure*}

\begin{figure}[!bp]
    \centering
	\begin{minipage}[t]{0.49\linewidth}
        \centering
    	\includegraphics[width=1\linewidth]{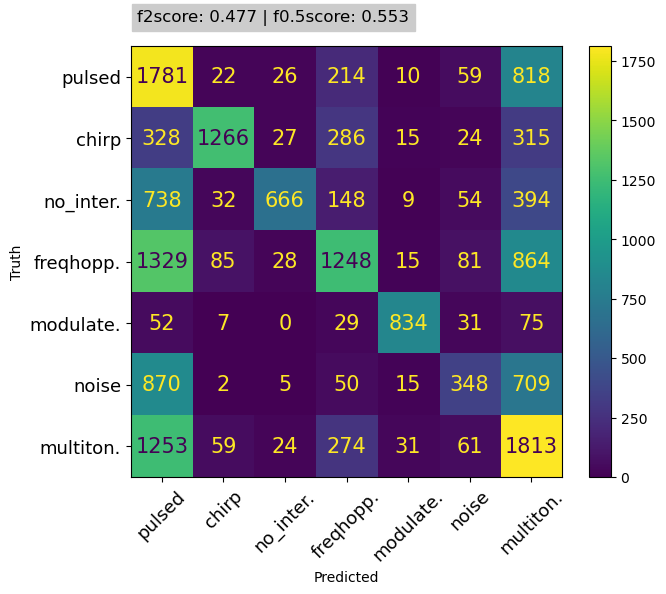}
        \subcaption{Spectral domain.}
    \end{minipage}
    \hfill
	\begin{minipage}[t]{0.49\linewidth}
        \centering
    	\includegraphics[width=1\linewidth]{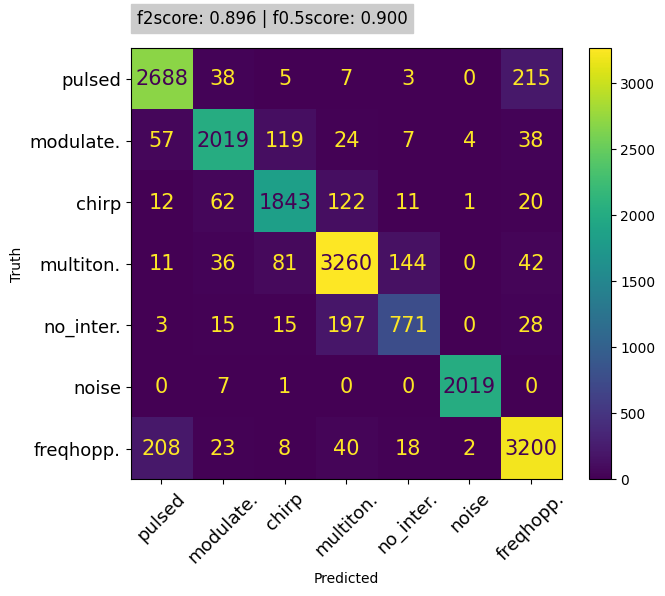}
        \subcaption{Spectral and temporal domains combined.}
    \end{minipage}    
    \caption{Classification accuracy for the best AE architectures.}
    \label{spectral-mixed-ae}
\end{figure}

The quantized AE runs solely on the Edge TPU. To calculate its power consumption, we reference the DevBoard Mini Datasheet~\cite{DevBoardMiniDatasheet} (2 TOPS per watt) and real-world measurements from Libutti et al.~\cite{Libutti2020BenchmarkingPA}, which assess power usage of USB accelerators~\cite{USBAccelerator} with MLPerf~\cite{MLPerf}. Operating at maximum frequency, the Edge TPU consumes between 1.6\,W and 2.0\,W. The largest model considered is a VAE for IQ data, with three hidden layers (128 neurons each) and three convolutional layers, totaling over 100M parameters. This model requires approximately 11,500 times more operations per batch than the image classification model in~\cite{Libutti2020BenchmarkingPA}.

\subsection{Best Autoencoder Architectures}
\label{BestAutoencders}

The final decision is a balance of several factors. For example, the AE with three latent variables has nearly 100k parameters, occupying most of the memory, and requires close to 2M operations. In contrast, the AE with the fewest parameters (26k) uses 9 latent variables and involves around 520k operations. Ultimately, we select the AE offering the best compressibility (latent variables and memory size) while maintaining a good balance with downstream classification accuracy. We evaluate these architectures in the spectral domain (Section~\ref{AE_spectral_domain}), temporal domain (Section~\ref{AE_temporal_domain}), and combined domains (Section~\ref{AE_domains_combined}). Figure~\ref{fig:AE_best_architectures} presents the final selected AE architectures.

\vspace{+0.1cm}
\textbf{Spectral Domain.}
\label{AE_spectral_domain}
The AE with the best compressibility features the following architecture: two hidden layers, each with 128 neurons, and four latent variables, totaling 68k parameters and 1.3M operations per input. This results in a compression rate that is 33\% higher than that of the smallest AE, which has 26k parameters and 520k operations per input. The AE achieves an $\text{F}_{2}$-score of 0.678 and an $\text{F}_{0.5}$-score of 0.734 for downstream classification. In comparison, the best-performing AE, which includes 9 latent variables, achieves an $\text{F}_{2}$-score of 0.688 and an $\text{F}_{0.5}$-score of 0.745.

\vspace{+0.1cm}
\textbf{Temporal Domain.}
\label{AE_temporal_domain}
For the temporal domain, the AE with the best compressibility features the following architecture: two hidden layers, each with 64 neurons, and three latent variables. This AE has the fewest latent variables while achieving the best downstream classification accuracy, with 15k parameters and 292k operations per input. The AE attains an $\text{F}_{2}$-score of 0.853 and an $\text{F}_{0.5}$-score of 0.861.

\vspace{+0.1cm}
\textbf{Spectral and Temporal Domain.}
\label{AE_domains_combined}
Architecture of AE with the best compressibility on mixed spectral and temporal domain (see Fig.~\ref{spectral-mixed-ae}): 2 hidden layers, each with 128 neurons, and 6 latent variables, totaling 80k parameters and 1.6M operations. This AE achieves an $\text{F}_{2}$-score of 0.888 and an $\text{F}_{0.5}$-score of 0.894 for downstream classification.

\subsection{Results Disentanglement with Factorized VAE}
\label{label_factorVAE_results}

Figure~\ref{label_disentanglement_results} presents the hyperparameter results obtained using FactorVAE for disentanglement and training a compact model on reconstructed (small latent) embeddings across three different datasets. The dashed lines represent results for a latent dimensionality of 32, while the remaining lines correspond to a dimensionality of 128. The bold black line denotes the baseline ResNet18 model. The following conclusions can be drawn: (1) Performance close to the baseline is achieved using compressed data, indicating that the VAE effectively encodes essential information. (2) ResNet18 (11M parameters) demonstrates superior decoding performance compared to a convolutional layer (1M parameters), despite its larger size. (3) In the context of dense layers, ReLU activation outperforms batch normalization. (4) The reparameterization step, which involves random sampling from a normal distribution, negatively impacts performance. (5) The $\mu \log$ configuration shows a marginally better performance compared to $\mu$.
\section{Conclusion}
\label{sec:summary}

Our experiments effectively demonstrate the capability of GenAI in achieving high classification accuracies, with $\text{F}_{2}$-scores reaching 0.988 for data representative of real-world scenarios. The application of a variational AE (VAE) on the Google Edge TPU not only enabled efficient data reconstruction -- achieving a compression rate of 97.6\%, reducing the input from 253 values per second to just six -- but also yielded significant energy savings, lowering power consumption from 1,203\,mWh to 130\,mWh. Moreover, the conversion pipeline for transitioning PyTorch models to TensorFlow Lite proved effective across various VAE architectures. The integration of both temporal and spectral domain data further demonstrated its potential for enhancing classification accuracy. These findings underscore the considerable promise of leveraging the Google Edge TPU in proximity to receiver hardware for improving GNSS interference detection and classification systems. Additionally, we demonstrated that disentanglement through factorized VAEs can successfully interpolate between latent variables of GNSS snapshots, a capability that can be leveraged for data generation and ML model augmentation.

In the spectral domain, the classification pipeline achieved an $\text{F}_{2}$-score of 0.937 on unknown test data, which dropped to 0.916 after reconstruction. Conversely, the temporal domain showed superior results, with an $\text{F}_{2}$-score of 0.995 on unknown test data, decreasing to 0.983 following reconstruction. The mixed temporal and spectral domains exhibited slight improvements, attaining an $\text{F}_{2}$-score of 0.989 on unknown test data and decreasing marginally to 0.988 after reconstruction. Applying VAE models achieved substantial compression rates, particularly in the combined temporal and spectral domains, where data was reduced from 253 values per second to just 6 values (mean and variance), yielding a compression rate of approximately 97.6\%. This significant compression minimizes data transmission requirements and enhances system efficiency. Notably, GenAI reconstructed interference-aware GNSS signals with only a 1\% loss in $\text{F}_{2}$, demonstrating that VAEs can effectively compress GNSS signals without significantly degrading classification performance, thus supporting practical applications in efficient signal processing.

\bibliography{ION_PLANS}
\bibliographystyle{IEEEtran}

\end{document}